\documentclass[journal]{IEEEtran}

\usepackage{cite}
\usepackage{amsmath,amssymb}
\usepackage{graphicx}
\usepackage{booktabs,array,multirow,adjustbox}
\usepackage{xcolor}
\usepackage{microtype}
\usepackage{fvextra}
\usepackage{url}
\usepackage{placeins}
\usepackage[hidelinks]{hyperref}

\graphicspath{{media/}}
\DefineVerbatimEnvironment{SSAVAlgorithm}{Verbatim}{fontsize=\scriptsize,breaklines=true,breakanywhere=true,frame=single,xleftmargin=0.4em,xrightmargin=0.4em}

\title{Semantic-Spatial Agreement Verification for Mitigating Object Hallucination in Multimodal Large Language Models}

\author{Ziheng Ren, Qian Gao, Jun Fan, Guohui Ding, Zhenyu Yang, and Yuteng Xiao%
\thanks{This work was supported by the Shandong Provincial Natural Science Foundation under Grants ZR2022MF333 and ZR2024QF053; the Pilot Project for Integrated Innovation of Science, Education, and Industry of Qilu University of Technology (Shandong Academy of Sciences) under Grant 2026ZDCX01; and the Key Laboratory of Computing Power Network and Information Security, Ministry of Education, under Grants 2023ZD028 and 2024PY025.}%
\thanks{Ziheng Ren, Qian Gao, Zhenyu Yang, and Yuteng Xiao are with the Key Laboratory of Computing Power Network and Information Security, Ministry of Education, Shandong Computer Science Center (National Supercomputer Center in Jinan), Qilu University of Technology (Shandong Academy of Sciences), Jinan 250014, China; the Shandong Engineering Research Center of Big Data Applied Technology, Faculty of Computer Science and Technology, Qilu University of Technology (Shandong Academy of Sciences), Jinan 250353, China; and the Shandong Provincial Key Laboratory of Industrial Network and Information System Security, Shandong Fundamental Research Center for Computer Science, Jinan 250014, China. E-mail: a49155464@163.com (Ziheng Ren); gq@qlu.edu.cn (Qian Gao, corresponding author).}%
\thanks{Jun Fan is with China Telecom Digital Intelligence Technology Co., Ltd., Jinan 250101, China.}%
\thanks{Guohui Ding is with Shenyang Aerospace University, Shenyang, China.}%
\thanks{Source code is publicly available at \url{https://github.com/zihengren/SSAV}.}}

\begin{document}
\maketitle

\begin{abstract}
Multimodal large language models generate natural-language responses from visual inputs, yet may mention objects absent from an image. In medication assistance, accessible perception, and environmental decision-making, such hallucinations can create real-world safety risks. We propose Semantic-Spatial Agreement Verification (SSAV), a training-free method for verifying object claims. A visually grounded claim should remain stable across semantically equivalent queries and repeatedly localize to the same image region. SSAV aggregates multiple prompts to estimate semantic support and reduce sensitivity to query wording. Query-Induced Regional Verification (QIRV) combines cross-query region persistence, spatial overlap, and relative candidate dominance to identify isolated high responses and dispersed localizations. A geometric mean fuses semantic and spatial evidence, lowering the verification score when either branch lacks support. Experiments on three base models and multiple evaluation protocols show that SSAV effectively mitigates object hallucination. On LLaVA-1.5-7B, accuracy averaged across COCO, A-OKVQA, and GQA improves by 1.81 and 3.17 percentage points under POPE Popular and Adversarial, respectively, while CHAIRs decreases from 49.40\% to 32.80\%. These results show that cross-query semantic stability and regional consistency provide interpretable external visual evidence for object claims.
\end{abstract}


\section{Introduction}

\IEEEPARstart{M}{ultimodal} large language models (MLLMs) typically consist of a visual encoder, a cross-modal connector, and an autoregressive language model. Visual encoders commonly adopt CLIP-family Vision Transformers \cite{radfordLearningTransferableVisual2021,dosovitskiy2021image} to encode an input image into visual tokens. Cross-modal connectors then project, compress, and align visual features with the representation space of the language model through linear projections, multilayer perceptrons, or Q-Former architectures \cite{li2023blip2}. For example, LLaVA connects a CLIP visual encoder to a language model through a projection module \cite{NEURIPS2023_6dcf277e}, whereas BLIP-2, InstructBLIP, and MiniGPT-4 use a Q-Former to extract and compress visual information relevant to language generation \cite{li2023blip2,daiInstructBLIPGeneralpurposeVisionLanguage2023,zhuMiniGPT4EnhancingVisionLanguage2023}. These architectures support image-based question answering, image captioning, and multi-turn interaction, yet responses can still mention objects absent from the image after visual encoding, cross-modal alignment, and autoregressive generation. Such object hallucinations make it difficult for users to judge whether a response is trustworthy. A visually impaired user may rely on a visual assistant to distinguish medicines with similar appearances but different indications or dosages; confusing one medicine with another, or falsely claiming that the target medicine is present, may cause medication errors. Likewise, missing an obstacle or inventing safety equipment in accessible navigation, industrial inspection, or emergency response may prompt unsafe actions \cite{tang2025blindHallucination}. Object hallucination is therefore not ordinary linguistic noise but a safety problem that can be amplified along an error chain from perception to judgment and action. Mitigating hallucination is consequently a prerequisite for deploying MLLMs in high-reliability settings.

Existing research improves visual faithfulness through two main routes: training-stage alignment and inference-time intervention. Training-stage methods use hallucination-aware visual instruction data, hard negatives, or fine-grained image descriptions to supervise all or part of a model. Some further introduce vision--language alignment losses, contrastive objectives, reinforcement learning from human feedback, or direct preference optimization \cite{liu2024mitigating,li2024vlfeedback}. These approaches reduce overreliance on linguistic co-occurrence patterns at the data and parameter levels, but the cost of high-quality data construction, preference annotation, and model updating limits rapid transfer. Inference-time methods instead reweight or select visual tokens, intervene in attention or hidden states, calibrate candidate distributions, perform contrastive decoding, or apply retrospective self-correction and post-generation revision \cite{leng2024mitigating,huangOPERAAlleviatingHallucination2024,zhang2025degf,yin2023woodpecker}. They avoid retraining but often require access to specific network layers, attention structures, or autoregressive decoding interfaces; some also rely on additional sampling, backtracking, or multiple generation rounds. More importantly, internal confidence or attention strength does not inherently establish that an object is present. Once a model is influenced by linguistic priors, redistributing information within the same generation system may preserve biases that share the same source as the original error.

A complementary approach is to verify an object claim with an independent visual model. Open-vocabulary object detectors accept textual queries and return matched local regions with corresponding scores \cite{minderer2022owlvit,liu2024groundingdino}, thereby providing a natural interface for object-level external verification. Compared with direct intervention in the internal state of an MLLM, this route reconnects a language response to localizable image regions and makes the visual basis of a judgment easier to interpret. However, an open-vocabulary detector is not an infallible fact judge. Different expressions of the same object can change text--region matching scores, making a decision based on one query and a fixed threshold sensitive to wording. Background textures, local contours, co-occurring objects, or excessively broad candidate boxes can also yield isolated high responses. Even when the maximum score exceeds a threshold, the response may not correspond to a stable object instance. A single-query maximum therefore represents only one local matching result and cannot establish whether an object claim is repeatedly supported by the same visual evidence.

We revisit external object verification from the perspective of cross-query consistency. If an object is present, semantically equivalent queries may produce different absolute scores, but they should provide reproducible semantic support overall and repeatedly localize to the same or highly overlapping regions. Conversely, a spurious response triggered by particular wording, background texture, or an incidental local feature is more likely to vary across prompts or disperse across unrelated regions. Object-presence evidence should therefore not be compressed into one scalar from a single detection. It should jointly capture semantic stability across prompts and spatial consistency across query-induced localizations. These two dimensions respectively ask whether different expressions consistently support the concept and whether such support originates from the same candidate instance, forming a stricter visual-verification condition than a single-query threshold.

Based on this insight, we propose Semantic-Spatial Agreement Verification (SSAV). For each object claim, SSAV constructs four semantically equivalent queries and retains high-scoring candidates returned for each query by an open-vocabulary detector. The semantic branch aggregates the strongest matching scores across queries to reduce response fluctuations caused by a single template. The spatial branch introduces Query-Induced Regional Verification (QIRV), which treats the preferred box of each query as a node, connects boxes according to intersection over union, and jointly measures candidate persistence, spatial overlap, and the dominance of the preferred candidate over the runner-up within the largest connected component. Semantic and spatial evidence are fused by a geometric mean so that a clear deficiency in either dimension lowers the final verification score. For object-existence question answering, SSAV calibrates the original Yes--No logit margin with external evidence rather than replacing the MLLM prediction. For open-ended image description, the same evidence verifies object claims in the generated response, allowing a common verification mechanism to support different output forms.

The main contributions are as follows:
\begin{itemize}
\item We decompose unreliable single-query external object verification into prompt-induced variation in semantic responses and inconsistency among candidate regions, and formulate cross-query semantic-spatial agreement as evidence for visual support.
\item We develop SSAV and its spatial branch QIRV, which jointly verify whether different queries point to the same object instance through candidate persistence, spatial overlap, and candidate dominance, complementing multi-prompt semantic evidence.
\end{itemize}

We evaluate SSAV on three MLLMs with different architectures, LLaVA-1.5-7B, mPLUG-Owl2, and MiniGPT-4 Vicuna-13B. Closed-set object-existence judgments are evaluated on the COCO, A-OKVQA, and GQA sources of POPE under Random, Popular, and Adversarial sampling and on paired Yes--No decisions in MME-Existence. Sentence-level and instance-level object hallucinations in open-ended descriptions are evaluated with CHAIR on 500 COCO images. Core-component ablations, QIRV evidence-correspondence shuffling, fusion-strength sensitivity, and qualitative cases further examine the mechanisms of multi-prompt semantic support and spatial verification. Across the three base models, SSAV improves object-existence judgments and reduces both CHAIRs and CHAIRi. For LLaVA-1.5-7B, average POPE Accuracy improves by 1.81 and 3.17 percentage points under Popular and Adversarial sampling, respectively, while CHAIRs and CHAIRi decrease from 49.40\% and 13.82\% to 32.80\% and 8.22\%. These cross-model, cross-source, cross-task, and mechanism-level results show that SSAV effectively mitigates object hallucination in MLLMs.

\section{Related Work}

\subsection{Object Hallucination and Mitigation in MLLMs}

Object hallucination denotes an MLLM claim unsupported by the input image, including mentions of absent objects, incorrect categories, or attributes, counts, actions, and relations generated around nonexistent objects \cite{rohrbachObjectHallucinationImage2018,liEvaluatingObjectHallucination2023}. Unlike factual hallucination in text-only settings, it can be checked directly against the image and reflects a mismatch between model output and visual evidence. Object hallucination is commonly associated with weakened transmission of fine-grained visual information and strong linguistic priors. Visual projection, compression, and insufficient cross-modal alignment may weaken object evidence, while previously generated text, frequent co-occurrences, and linguistic momentum increasingly influence autoregressive decoding. The resulting response can remain fluent despite lacking stable visual support \cite{leng2024mitigating,seo2025epistemic}.

Existing mitigation methods use either training-stage optimization or training-free inference intervention. Training-stage approaches employ hallucination-aware data, hard negatives, regional descriptions, alignment or contrastive objectives, and human-feedback or preference optimization, but require additional annotation, model updates, and computation. At inference time, visual contrastive decoding compares perturbed visual inputs or generation distributions \cite{leng2024mitigating}; DeCo and DeGF apply distribution correction or generative feedback \cite{huang2024mllm,zhang2025degf}; PAI, CCA, and VASparse redistribute visual tokens or attention \cite{liu2024paying,xing2024cca,zhuang2025vasparse}; KVSmooth smooths key--value caches \cite{jiang2026kvsmooth}; and CIPHER and TruthPrInt intervene along hallucination- or truthfulness-related directions \cite{dastmalchi2026cipher,duan2025truthprint}. Other approaches construct contrastive views, compare instruction-conditioned distributions, introspect decoding states, or revise generated descriptions \cite{chen2024halc,park2025convis,wang2024mitigating,huo2025self,zhou2024lure}. Together, these methods address underused visual information through decoding distributions, attention structures, and latent representations.

Training-stage optimization depends on data and model updates, whereas internal inference intervention usually requires access to specific attention layers, hidden states, visual tokens, or next-token distributions; some methods also perform multiple forward passes, repeated sampling, or backtracking. Moreover, internal confidence, attention magnitude, or a latent direction characterizes the model's own response state and does not inherently establish that an object is present, while overly strong intervention may also suppress visually grounded objects. SSAV instead leaves the parameters, internal representations, and autoregressive decoding of the MLLM unchanged. After an object claim is formed, it calls an independent visual model to provide localizable evidence and calibrates the original decision margin in closed-set tasks. The external model does not replace the MLLM; it verifies the object claim with independent visual evidence, reducing dependence on a particular internal network structure.

\subsection{External Visual Verification and Post-Generation Revision}

External visual verification separates factual checking from the original generator. A typical pipeline extracts objects, attributes, and relations from an answer or caption; invokes an object detector, visual question-answering model, or other visual tool; and then deletes, rewrites, or regenerates unsupported content. Compared with self-checking based only on language probabilities or generation confidence, external verification reconnects textual claims to image content and provides a more interpretable visual basis. Prior work has used object-wise questions, multi-round visual question answering, region cropping, and local re-identification. Other approaches use text-to-image models to construct external feedback that guides correction during generation \cite{yin2023woodpecker,zhang2025degf}.

External tools do not guarantee reliable factual decisions. Multi-round verification can inherit verifier bias and propagate early errors, while a generative visual question-answering verifier may itself hallucinate. Single-threshold post-processing can mistake detection confidence for object-presence probability and ignore query wording, category coverage, and candidate-region quality. The central issue is therefore how to construct stable, repeatable, and interpretable object-presence evidence.

SSAV verifies object claims with independent visual evidence without multi-round MLLM reasoning or rule-based deletion of low-confidence objects. It converts each claim into semantically equivalent queries and jointly analyzes detector scores and candidate regions. The semantic branch evaluates cross-query conceptual support, whereas the spatial branch determines whether responses repeatedly localize to the same region. External evidence is therefore derived from consistency across queries rather than a single model answer or local match.

\subsection{Open-Vocabulary Object Detection and Regional Consistency}

Open-vocabulary object detection jointly encodes image regions and text queries, directly aligning category names or natural-language phrases with candidate boxes \cite{gu2022vild,li2022glip,minderer2022owlvit,liu2024groundingdino}. Representative developments further learn detection-specific prompts or region--text representations, expand detector vocabularies with image-level supervision, reuse frozen vision--language models, and scale open-vocabulary self-training \cite{duLearningPromptOpenVocabulary2022,zhong2022regionclip,zhou2022detic,kuo2023fvlm,minderer2023scaling,wu2024openvocabulary}. Unlike a detector with a fixed category set, it can respond to open-form object claims without retraining a classification head for each verification target, making it a natural interface for object-level verification of MLLM outputs. A text--region matching score describes how strongly a candidate region responds to the query, while the box localizes the spatial origin of that response. Open-vocabulary detection, visual grounding, and region cropping have consequently been used to check generated content and provide regional visual evidence \cite{yin2023woodpecker}.

Detection scores are generally intended for candidate ranking rather than as uniformly calibrated object-presence probabilities. Querying the same object with a class name, a noun phrase, or a scene-conditioned expression can substantially change the text embedding and matching score. Background texture, neighboring objects, and overly broad candidate boxes can also receive high responses. A single-query maximum therefore establishes only a strong match between one expression and one region; it does not show that the object claim remains supported as the expression changes. If the strongest responses to different queries fall in nonoverlapping regions, even individually high scores are difficult to interpret as repeated evidence for the same visual instance.

Regional consistency provides a complementary signal for separating stable instance responses from incidental local activations. When semantically equivalent queries repeatedly localize to the same object, their candidate boxes should form a region cluster with high query coverage and substantial internal overlap. Responses triggered by background content or query-specific spurious matches are more likely to disperse across locations. The advantage of the highest-scoring candidate over the runner-up further indicates whether a query produces a clear local dominant response. QIRV therefore combines region persistence, spatial consistency, and within-query candidate dominance and fuses this spatial evidence with multi-prompt semantic support. Rather than asking whether one box exceeds a fixed threshold, it asks whether equivalent queries agree in both response strength and spatial instance.

\section{Method}

\subsection{Problem Formulation and Method Overview}

Given an input image $I$, a question $q$, and a frozen multimodal large language model $M$, the base model generates $y=M(I,q)$. We focus on claims $c$ in the response that can be reduced to whether an object exists, and seek to determine whether each claim has sufficient and stable image evidence without updating the base model. In closed-set object-existence question answering, the target object is obtained directly from the question. In open-ended image description, a set of object claims $\mathcal{C}(y)$ is extracted from the generated text using a predefined category vocabulary and synonym mapping. Claim extraction identifies verification targets only and does not participate in evidence scoring.

To obtain visual evidence independent of the internal state of the base model, we use a frozen open-vocabulary detector $D$ \cite{minderer2022owlvit} as an auxiliary verifier. Given image $I$ and an object query, $D$ returns candidate regions and their text--region matching scores. The external detector neither replaces the base model nor turns a single detection maximum directly into an object-presence probability. Instead, it provides comparable regional responses that allow a claim to be evaluated along two dimensions: whether support persists across expressions and whether different expressions point to the same instance.

\begin{figure}[t]
\centering
\includegraphics[width=\columnwidth]{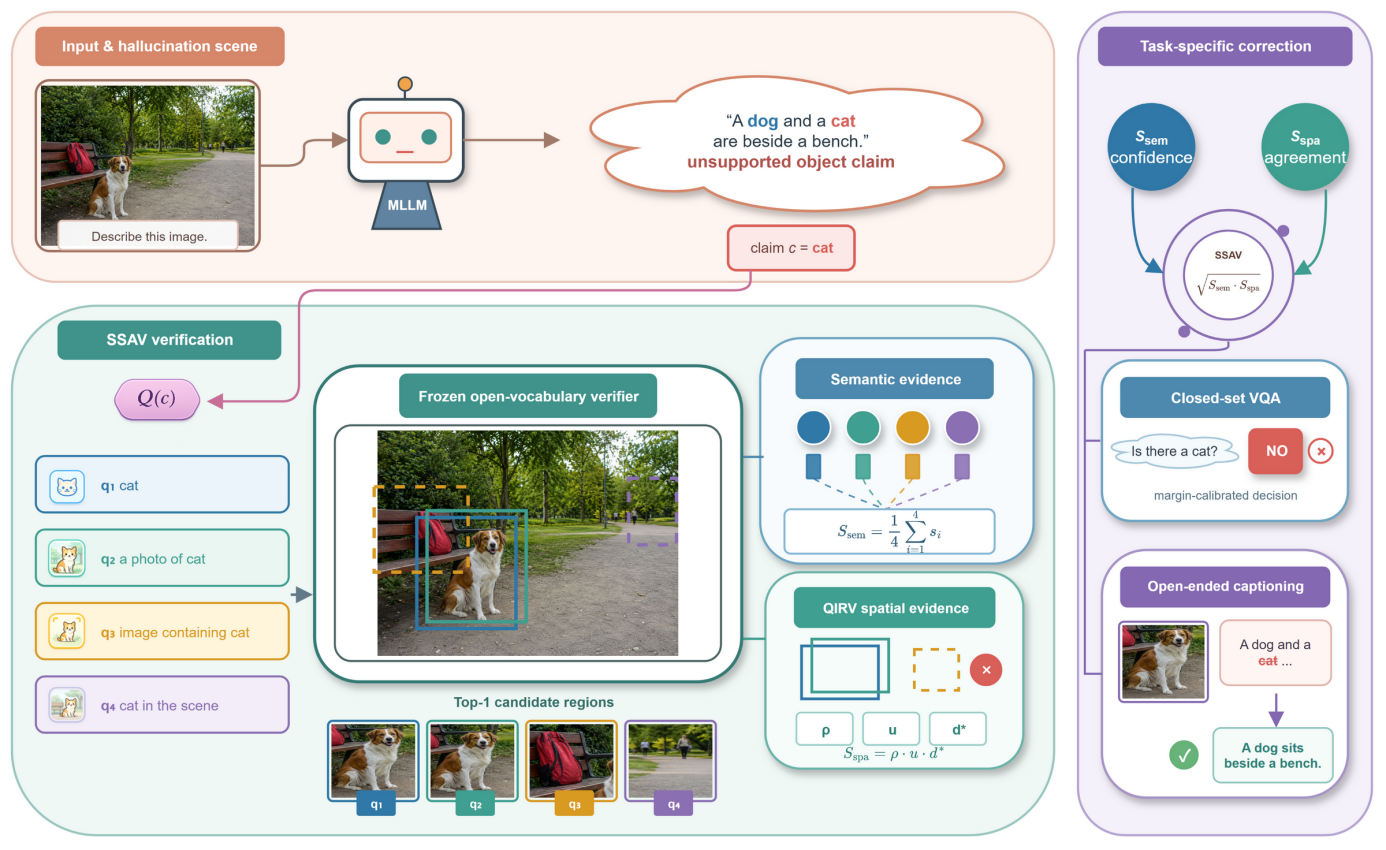}
\caption{Overview of Semantic-Spatial Agreement Verification (SSAV). An object claim is extracted from the MLLM response and converted into four semantically equivalent queries. A frozen open-vocabulary verifier returns candidate regions for each query. SSAV combines multi-query semantic support with QIRV spatial evidence, which measures region persistence, spatial overlap, and candidate dominance. The fused evidence calibrates the Yes--No margin in closed-set VQA or filters unsupported object claims in open-ended captioning.}
\label{fig:overview}
\end{figure}

The SSAV pipeline is shown in Fig.~\ref{fig:overview}. First, the multi-prompt semantic-support branch constructs fixed semantically equivalent queries for claim $c$ and aggregates the highest matching score from each query, reducing variation caused by one wording. Second, Query-Induced Regional Verification (QIRV) builds a region-relation graph from the preferred boxes produced by different queries and tests whether the responses bind to the same visual instance through candidate persistence, spatial consistency, and within-query candidate dominance. The two forms of evidence are fused by a geometric mean into an object-level verification score $S_{\mathrm{ssav}}$. Closed-set object-existence question answering and open-ended image description share this evidence-extraction process but use different decision rules: the former calibrates the base model's Yes--No logit margin, whereas the latter filters unsupported object claims with a claim-level threshold.

\begin{figure}[t]
\centering
\includegraphics[width=0.86\columnwidth]{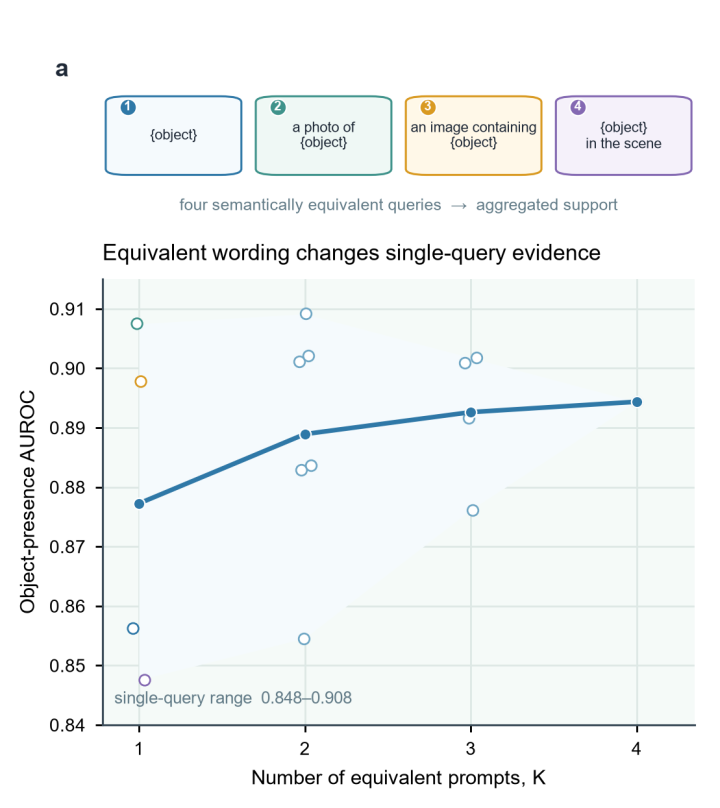}
\vspace{2pt}

\makebox[\columnwidth][l]{\hspace*{0.035\columnwidth}\includegraphics[width=0.86\columnwidth]{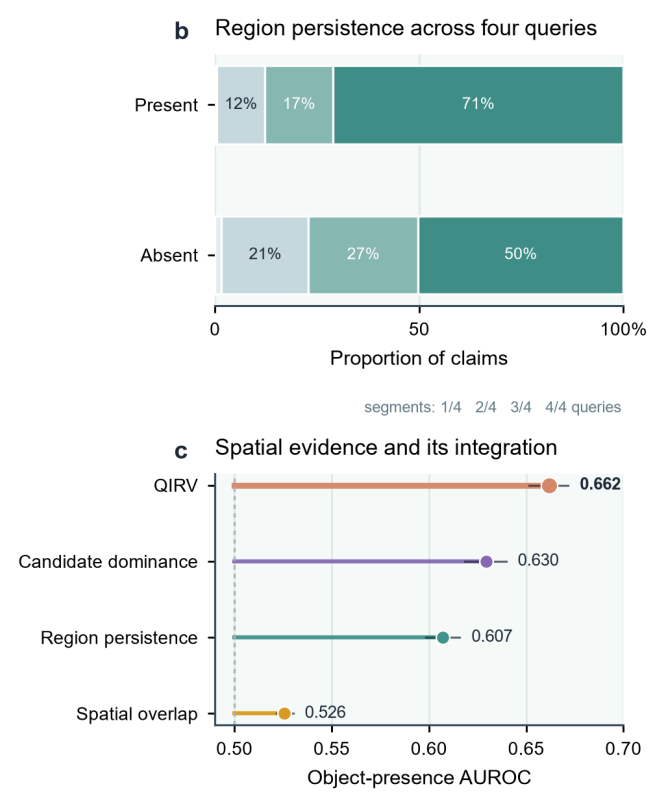}}
\caption{Empirical motivation for semantic-spatial verification in SSAV. (a) Object-presence AUROC across different numbers of semantically equivalent queries; the line and shaded region show the mean and range across prompt combinations. (b) Distribution of the largest region-cluster size for present- and absent-object claims. (c) Object-presence AUROC of QIRV and its three spatial components, with image-clustered 95\% bootstrap confidence intervals.}
\label{fig:motivation}
\end{figure}

Figure~\ref{fig:motivation} presents the empirical observations motivating the method. Aggregating equivalent queries improves the stability of object-presence evidence; present-object claims more often localize repeatedly to the same region across queries; and QIRV discriminates object presence better than any individual spatial component. These observations motivate the multi-prompt semantic-support and spatial-instance verification designs described below.

\subsection{Multi-Prompt Semantic Support}

The matching score of an open-vocabulary detector is affected by query wording. Even when several text queries express the same object concept, they may activate regional responses of different strengths. Templates can also introduce different linguistic and contextual biases, making a high response to one query insufficient as stable visual support. Verification based only on a class-name query can therefore depend excessively on one expression \cite{zhangLowRankPromptAdaptation}. To reduce this prompt sensitivity, we construct a query set $\mathcal{P}(c)$ containing $K=4$ fixed templates for each object claim $c$:
\begin{equation}
\begin{aligned}
\mathcal{P}(c)=\{&c,\ \text{a photo of }c,\ \text{an image containing }c,\\
&c\ \text{in the scene}\}.
\end{aligned}
\end{equation}
All templates are fixed before testing and are not selected for a dataset, base model, or individual sample during evaluation. To prevent multiple text queries from interacting within one detector call, each query is submitted independently. For the $j$th query $p_j\in\mathcal{P}(c)$, the detector retains the top $N$ candidate regions in descending order of matching score:
\begin{equation}
D(I,p_j)=\left\{\left(b_j^{(n)},s_j^{(n)}\right)\right\}_{n=1}^{N},\quad
s_j^{(1)}\ge s_j^{(2)}\ge\cdots\ge s_j^{(N)}.
\end{equation}
Here, $b_j^{(n)}$ is the $n$th candidate box and $s_j^{(n)}\in[0,1]$ is its text--region matching score. The semantic branch averages the preferred candidate score from each query to obtain multi-prompt semantic support:
\begin{equation}
S_{\mathrm{sem}}(c)=\frac{1}{K}\sum_{j=1}^{K}s_j^{(1)}.
\end{equation}
$S_{\mathrm{sem}}$ does not require identical scores across queries. Instead, cross-query aggregation reduces the influence of an anomalous response from a single template. Support is retained when a claim receives strong responses under several equivalent expressions, whereas a score triggered only by particular wording is attenuated after aggregation. Semantic support, however, captures only cross-query response strength and cannot determine whether the responses originate from the same object, different objects, or a broad background region. We therefore verify spatial instance consistency using the relationships among candidate boxes.

\subsection{Query-Induced Regional Verification}

A high semantic score does not necessarily provide reliable instance-level evidence. Different queries may yield high scores at different image locations or jointly activate a broad background region, and averaging their scores ignores these localization differences. QIRV therefore treats the preferred candidate from each query as a query-induced region hypothesis and tests whether the resulting regions consistently correspond to the same spatial instance.

Specifically, QIRV takes the $K$ preferred boxes $\{b_j^{(1)}\}_{j=1}^{K}$ as nodes and constructs an undirected graph $G=(V,E)$. For any two queries $i$ and $j$, an edge is added if the intersection over union of their preferred boxes is no smaller than threshold $\eta$:
\begin{equation}
(i,j)\in E \iff \operatorname{IoU}\!\left(b_i^{(1)},b_j^{(1)}\right)\ge\eta.
\end{equation}
We set $\eta=0.5$ and denote the largest connected component of $G$ by $\mathcal{C}^{\star}$. Connected components organize regional responses from different queries into candidate instance clusters, and the largest component represents the spatial instance supported by the greatest number of queries. If several queries repeatedly point to the same object, their preferred boxes should concentrate in one connected component and exhibit substantial overlap. QIRV consequently computes candidate persistence, spatial consistency, and within-query candidate dominance in sequence.

Candidate persistence measures the proportion of queries covered by the largest connected component:
\begin{equation}
\rho=\frac{|\mathcal{C}^{\star}|}{K}.
\end{equation}
When all queries point to mutually connected regions, $\rho=1$; when only a few queries share a spatial response, $\rho$ decreases. Thus, $\rho$ measures the persistence of a candidate instance under changes in query wording rather than the confidence of a single box.

Candidate persistence counts the queries covered by the component but does not capture its geometric compactness. When $|\mathcal{C}^{\star}|>1$, we define spatial consistency $u$ as the mean pairwise IoU among all boxes in the largest connected component:
\begin{equation}
u=\frac{2}{|\mathcal{C}^{\star}|(|\mathcal{C}^{\star}|-1)}
\sum_{\substack{i<j\\i,j\in\mathcal{C}^{\star}}}
\operatorname{IoU}\!\left(b_i^{(1)},b_j^{(1)}\right).
\end{equation}
If the largest connected component contains only one node, no box pair is available. We set $u=1/K$ to retain limited support for an isolated response while preventing it from being interpreted as sufficient spatial consistency. Hence, $\rho$ and $u$ respectively describe how many queries support one candidate instance and how tightly those supports align in space.

Even when preferred boxes from several queries overlap, the detector may assign similar scores to multiple regions within each query. The preferred candidate then lacks a clear advantage and remains spatially ambiguous. To quantify this local competition, QIRV compares the preferred and runner-up candidates for each query. Let their scores be $s_j^{(1)}$ and $s_j^{(2)}$; within-query candidate dominance is defined as
\begin{equation}
d_j=\sigma\!\left(\operatorname{logit}\!\left(s_j^{(1)}\right)-
\operatorname{logit}\!\left(s_j^{(2)}\right)\right),
\end{equation}
where $\sigma(\cdot)$ is the sigmoid function. Detection scores are clipped to $[\varepsilon,1-\varepsilon]$ before the logit to avoid numerical overflow. $d_j$ increases when the preferred candidate clearly exceeds the runner-up and approaches a neutral level when their scores are similar. Subsequent aggregation uses only nodes in $\mathcal{C}^{\star}$ and the corresponding $d_j$, keeping within-query competition aligned with the selected instance cluster.

Combining these factors, QIRV spatial evidence is
\begin{equation}
S_{\mathrm{spa}}(c)=\rho u\,\frac{1}{|\mathcal{C}^{\star}|}
\sum_{j\in\mathcal{C}^{\star}}d_j.
\end{equation}
This product jointly captures cross-query instance coverage, geometric overlap within the component, and relative within-query dominance. An isolated high score is jointly limited by $\rho$ and the single-node value of $u$; dispersed localization reduces $u$; and ambiguity between the preferred and runner-up candidates reduces $d_j$. QIRV does not predict the category again. Instead, it uses the candidate structure already produced by the detector to test whether a category response binds stably to a specific visual instance.

\subsection{Semantic-Spatial Evidence Fusion}

Multi-prompt semantic support measures whether an object concept repeatedly receives strong responses under equivalent expressions, whereas QIRV measures whether those responses point to a stable spatial instance. The two forms of evidence correspond to response strength and instance consistency; a deficiency in either may indicate inadequate visual support. Given this complementarity, we use their geometric mean as the final verification score:
\begin{equation}
S_{\mathrm{ssav}}(c)=\sqrt{S_{\mathrm{sem}}(c)S_{\mathrm{spa}}(c)}.
\end{equation}
The geometric mean treats the branches symmetrically and lets the lower value directly constrain the result. If queries receive high scores but localize to dispersed regions, $S_{\mathrm{spa}}$ reduces the fused score. If boxes overlap strongly but the overall matching strength is weak, $S_{\mathrm{sem}}$ likewise limits the result. A high $S_{\mathrm{ssav}}$ therefore requires both cross-query semantic support and spatial-instance support. The fusion introduces no learnable parameters, and the resulting object-level evidence is passed to the corresponding task-inference procedure.

\subsection{Task Inference With SSAV Evidence}

SSAV produces a unified visual-evidence score $S_{\mathrm{ssav}}$ for each object claim. To adapt this evidence to different output forms, we calibrate the base model's Yes--No logit margin in closed-set object-existence question answering and filter object claims from generated text in open-ended image description. Both tasks share the same evidence-extraction process and differ only in how the evidence is used during inference.

\subsubsection{Closed-Set Object-Existence Question Answering}

Let the logits assigned by the base model to the Yes and No answer tokens be $\ell_{\mathrm{yes}}$ and $\ell_{\mathrm{no}}$. The original decision margin is
\begin{equation}
m_{\mathrm{base}}=\ell_{\mathrm{yes}}-\ell_{\mathrm{no}}.
\end{equation}
SSAV does not overwrite the base prediction with the external detector output. Object-level evidence instead calibrates the original margin:
\begin{equation}
m_{\mathrm{ssav}}=m_{\mathrm{base}}+\lambda_{\mathrm{ssav}}
\left(S_{\mathrm{ssav}}-\tau_{\mathrm{ssav}}\right),\quad
\widehat{y}=\mathbb{I}\!\left[m_{\mathrm{ssav}}\ge0\right].
\end{equation}
Here, $\lambda_{\mathrm{ssav}}\ge0$ controls the strength of external-evidence calibration, and $\tau_{\mathrm{ssav}}$ is the center threshold of the external evidence. When $S_{\mathrm{ssav}}>\tau_{\mathrm{ssav}}$, the calibration term favors Yes; when $S_{\mathrm{ssav}}<\tau_{\mathrm{ssav}}$, it favors No. The final decision jointly depends on the original base-model margin and the strength of the external evidence.

\subsubsection{Open-Ended Image Description}

Given the base-model output $y$, we first extract the object-claim set $\mathcal{C}(y)$ and compute $S_{\mathrm{ssav}}(c)$ for every claim. Because generated text provides no Yes--No margin corresponding to an individual claim, a frozen claim-level threshold $\tau_{\mathrm{cap}}$ partitions claims into retained and rejected sets:
\begin{align}
\mathcal{C}_{\mathrm{keep}}&=\{c\in\mathcal{C}(y)\mid S_{\mathrm{ssav}}(c)\ge\tau_{\mathrm{cap}}\},\\
\mathcal{C}_{\mathrm{reject}}&=\mathcal{C}(y)\setminus\mathcal{C}_{\mathrm{keep}}.
\end{align}
For claims in $\mathcal{C}_{\mathrm{reject}}$, deterministic editing rules remove the corresponding object phrase or neutralize it when direct deletion would disrupt sentence structure. Claims that are not rejected and all other text are left unchanged. This procedure revises only object claims already present in the generated text and does not trigger another generation by the base model. Object claims are identified using a task-defined category vocabulary and synonym mapping independently of subsequent SSAV scoring; the vocabulary can therefore be replaced for a different category space without changing the verification process.

Algorithm~1 summarizes the complete procedure for one object claim. For an open-ended description containing multiple claims, evidence extraction is repeated for each claim, after which the original text is edited once using all verification outcomes.

\begin{figure*}[t]
\small
\textbf{Algorithm 1: Semantic-Spatial Agreement Verification (SSAV)}
\begin{SSAVAlgorithm}
Input: image I, question q, frozen base model M, open-vocabulary detector D,
       fixed query-template set P, candidate count N, IoU threshold eta,
       and fixed task-specific decision parameters
Output: SSAV-calibrated or filtered response y'

1:  y <- M(I,q)
2:  Extract object claim c from q or y
3:  Construct K queries P(c) from c
4:  for each query p_j in P(c) do
5:      Independently evaluate D(I,p_j) and retain the top-N boxes and scores
6:      Record (b_j^(1),s_j^(1)) and the runner-up score s_j^(2)
7:  end for
8:  S_sem <- (1/K) sum_j s_j^(1)
9:  Build graph G from the IoU between preferred candidate boxes
10: Extract the largest component C* and compute rho, u, and {d_j | j in C*}
11: S_spa <- rho * u * (1/|C*|) sum_(j in C*) d_j
12: S_ssav <- sqrt(S_sem * S_spa)
13: if the task is closed-set Yes-No VQA then
14:     m_base <- ell_yes - ell_no
15:     m_ssav <- m_base + lambda_ssav(S_ssav - tau_ssav)
16:     y' <- I[m_ssav >= 0]
17: else
18:     Partition c as retained or rejected using tau_cap
19:     Edit the original caption deterministically to obtain y'
20: end if
21: return y'
\end{SSAVAlgorithm}
\end{figure*}

\section{Experiments}

\subsection{Experimental Setup}

\subsubsection{Base Models and Implementation Details}

We evaluate SSAV on three MLLMs with representative architectures and language backbones: LLaVA-1.5-7B, mPLUG-Owl2, and MiniGPT-4 Vicuna-13B \cite{liuImprovedBaselinesVisual2024,yeMPLUGOwl2RevolutionizingMultimodal2024,zhuMiniGPT4EnhancingVisionLanguage2023}. We use public implementations of all base models. During inference, SSAV obtains external visual evidence from a frozen OWL-ViT-base-patch32 \cite{minderer2022owlvit}; neither the base model nor the open-vocabulary detector is updated. The same query templates, candidate count, region-relation construction, and semantic-spatial fusion rule are used across all base models and remain fixed for every evaluation sample.

We compare SSAV with representative methods. Vanilla denotes the original base-model output without hallucination mitigation. VCD, OPERA, and DeCo suppress linguistic priors by modifying decoding \cite{leng2024mitigating,huangOPERAAlleviatingHallucination2024,huang2024mllm}. MemVR, ClearSight, and REVIS enhance visual representations through visual-information reinjection or hidden-state intervention \cite{zouLookTwiceYou2025,yin2025clearsight,wu2026revis}. CCA, VASparse, PAI, and MiddleLayer primarily redistribute visual tokens or attention \cite{xing2024cca,zhuang2025vasparse,liu2024paying,jiang2025devils}. PM enhances target perception through local image magnification \cite{mao2026perceptionMagnifier}. DeGF revises answers through generative feedback \cite{zhang2025degf}, and PTI intervenes in the multimodal KV cache during prefilling \cite{zhang2026pti}.

The hyperparameter $\lambda_{\mathrm{ssav}}$ is set to 3, 1, and 8 for LLaVA-1.5-7B, mPLUG-Owl2, and MiniGPT-4 Vicuna-13B, respectively. Unless otherwise stated, these settings are retained across all benchmark evaluations.

\subsubsection{Evaluation Benchmarks and Metrics}

We evaluate object hallucination in both closed-set object-existence question answering and open-ended image description. Closed-set evaluation uses POPE \cite{liEvaluatingObjectHallucination2023} and MME-Existence. POPE includes three data sources---COCO, A-OKVQA, and GQA \cite{linMicrosoftCOCOCommon2014,schwenkAOKVQABenchmarkVisual2022,hudsonGQANewDataset2019}---with Random, Popular, and Adversarial negative-sampling settings for each. We report Accuracy and F1. For each sampling setting, we take the arithmetic mean over the three data sources and separately summarize Random, Popular, and Adversarial. Random primarily reflects ordinary object-existence judgment, whereas Popular and Adversarial further test hallucination under frequent and confusable negative objects.

For MME-Existence, we use the official object-existence subtask \cite{fuMMEComprehensiveEvaluation2025} and report Accuracy, paired accuracy (Acc+), and their sum, the MME Score. The complete MME benchmark also contains counting, position, color, optical-character-recognition, and knowledge-reasoning tasks, whereas SSAV targets whether an object claim is supported by image content. We therefore use the Existence subtask, which directly matches the definition of object hallucination, and pair it with POPE as a closed-set evaluation without conflating changes in other abilities.

Open-ended image description is evaluated with CHAIR \cite{rohrbachObjectHallucinationImage2018}. All models generate captions for the same 500 COCO images with a maximum of 512 newly generated tokens. CHAIRs measures the proportion of sentences containing at least one hallucinated object, and CHAIRi measures the proportion of hallucinated instances among generated object mentions; lower values are better. Recall measures coverage of real objects in the generated descriptions, with higher values indicating better coverage.

\subsection{Experimental Results}

Using identical model versions and complete standard protocols, we compare SSAV with the original base models and representative hallucination-mitigation methods.

\begin{table*}[t]
\caption{Comparison with representative methods on POPE. For each sampling setting, Accuracy and F1 are averaged across COCO, A-OKVQA, and GQA. \textsuperscript{\dag} denotes results reported by subsequent studies using the same model backbone and evaluation protocol.}
\label{tab:pope}
\centering
\scriptsize
\setlength{\tabcolsep}{2.7pt}
\begin{tabular}{llcccccc}
\toprule
\multirow{2}{*}{\textbf{Model}} & \multirow{2}{*}{\textbf{Method}} & \multicolumn{2}{c}{\textbf{Random}} & \multicolumn{2}{c}{\textbf{Popular}} & \multicolumn{2}{c}{\textbf{Adversarial}} \\
& & \textbf{Acc $\uparrow$} & \textbf{F1 $\uparrow$} & \textbf{Acc $\uparrow$} & \textbf{F1 $\uparrow$} & \textbf{Acc $\uparrow$} & \textbf{F1 $\uparrow$} \\
\midrule
\multirow{10}{*}{LLaVA-1.5-7B}
& Vanilla & \textbf{90.33} & \textbf{90.27} & 84.76 & 85.62 & 79.91 & 81.85 \\
& VCD\textsuperscript{\dag} \cite{liu2025energy} & 82.31 & 83.83 & 76.59 & 79.82 & 71.20 & 76.14 \\
& HALC\textsuperscript{\dag} \cite{liu2025energy} & 87.00 & 87.81 & 80.31 & 82.76 & 72.68 & 77.49 \\
& Energy \cite{liu2025energy} & 88.49 & 87.90 & 84.33 & 84.30 & 79.98 & 80.79 \\
& DoLa\textsuperscript{\dag} \cite{an2025agla} & 84.78 & 84.19 & 79.75 & 80.61 & 76.32 & 76.16 \\
& OPERA\textsuperscript{\dag} \cite{an2025agla} & 87.53 & 86.45 & 84.21 & 83.50 & 80.88 & 80.69 \\
& AGLA \cite{an2025agla} & 88.54 & 87.71 & 85.14 & 84.68 & 81.13 & 81.36 \\
& ICD\textsuperscript{\dag} \cite{li2026cmac} & 87.82 & 86.73 & 84.72 & 83.94 & 80.98 & 80.78 \\
& PAI\textsuperscript{\dag} \cite{li2026cmac} & 88.32 & 88.01 & 84.69 & 85.61 & 79.18 & 80.38 \\
& \textbf{SSAV} & 88.54 & 87.49 & \textbf{86.57} & \textbf{85.66} & \textbf{83.08} & \textbf{82.60} \\
\midrule
\multirow{6}{*}{mPLUG-Owl2}
& Vanilla & 80.33 & 83.07 & 71.83 & 77.33 & 67.80 & 75.13 \\
& VCD\textsuperscript{\dag} \cite{liu2025energy} & 79.70 & 82.05 & 72.83 & 77.34 & 69.41 & 75.23 \\
& HALC\textsuperscript{\dag} \cite{liu2025energy} & 81.93 & 84.03 & 74.62 & 78.94 & 69.68 & 75.87 \\
& Energy \cite{liu2025energy} & 87.10 & 86.01 & 83.22 & 82.60 & 80.05 & 80.00 \\
& MVP \cite{qu2025mvp} & \textbf{90.14} & \textbf{90.10} & 82.49 & 83.32 & 77.70 & 79.60 \\
& \textbf{SSAV} & 88.10 & 87.44 & \textbf{84.58} & \textbf{84.34} & \textbf{80.78} & \textbf{81.24} \\
\midrule
\multirow{3}{*}{MiniGPT-4 Vicuna-13B}
& Vanilla & 76.06 & 80.80 & 70.50 & 75.09 & 64.76 & 71.66 \\
& Regular (POPE) \cite{liEvaluatingObjectHallucination2023} & 74.54 & 72.82 & 68.83 & 68.81 & 65.05 & 66.15 \\
& \textbf{SSAV} & \textbf{85.82} & \textbf{84.87} & \textbf{83.69} & \textbf{82.97} & \textbf{77.18} & \textbf{77.69} \\
\bottomrule
\end{tabular}
\end{table*}

Table~\ref{tab:pope} reports each sampling setting as the arithmetic mean over COCO, A-OKVQA, and GQA. On LLaVA-1.5-7B, SSAV raises Accuracy from 84.76\% to 86.57\% under Popular sampling and from 79.91\% to 83.08\% under Adversarial sampling; the corresponding F1 scores increase from 85.62\% to 85.66\% and from 81.85\% to 82.60\%. Random-sampling Accuracy and F1 change from 90.33\% and 90.27\% to 88.54\% and 87.49\%, respectively. Popular and Adversarial negatives contain frequent or strongly co-occurring confusable objects, for which the base model is more susceptible to linguistic priors. By requiring stable responses under equivalent queries and repeated localization to the same region, SSAV suppresses claims without consistent visual correspondences and yields larger gains in these difficult settings. In contrast, the strong Vanilla baseline under Random sampling leaves less room for correction; weakly supported true objects may be over-corrected, consistent with the recall trend in Fig.~\ref{fig:sensitivity}(b).

SSAV obtains the highest Accuracy and F1 within the LLaVA-1.5-7B and mPLUG-Owl2 groups under both Popular and Adversarial sampling. On mPLUG-Owl2, Adversarial Accuracy and F1 improve by 12.98 and 6.11 percentage points over Vanilla; the corresponding gains on MiniGPT-4 Vicuna-13B are 12.42 and 6.03 points. These larger gains indicate that independent semantic-spatial evidence is more beneficial when the base model distinguishes object presence from absence less reliably. Improvements across three language backbones in the difficult settings also show that the effect does not depend on a particular network layer or decoding interface.

\begin{table*}[t]
\caption{Same-protocol comparison with representative methods on MME-Existence. MME Score is the sum of Accuracy and Acc+. An em dash indicates that Accuracy and Acc+ were not reported separately. \textsuperscript{\dag} denotes results reported by subsequent studies using the same model backbone and evaluation protocol.}
\label{tab:mme}
\centering
\scriptsize
\setlength{\tabcolsep}{5pt}
\begin{tabular}{llccc}
\toprule
\textbf{Model} & \textbf{Method} & \textbf{Accuracy $\uparrow$} & \textbf{Acc+ $\uparrow$} & \textbf{MME Score $\uparrow$} \\
\midrule
\multirow{9}{*}{LLaVA-1.5-7B}
& Vanilla & 96.67 & 93.33 & 190.00 \\
& VCD\textsuperscript{\dag} \cite{liu2025energy} & --- & --- & 170.00 \\
& HALC\textsuperscript{\dag} \cite{liu2025energy} & --- & --- & 190.00 \\
& DoLa\textsuperscript{\dag} \cite{an2025agla} & --- & --- & 175.00 \\
& OPERA\textsuperscript{\dag} \cite{an2025agla} & --- & --- & 175.00 \\
& AGLA \cite{an2025agla} & --- & --- & 180.00 \\
& LURE\textsuperscript{\dag} \cite{wu2024logiccheckgpt} & 93.33 & 86.67 & 180.00 \\
& LogicCheckGPT \cite{wu2024logiccheckgpt} & 96.67 & 93.33 & 190.00 \\
& \textbf{SSAV} & \textbf{98.33} & \textbf{96.67} & \textbf{195.00} \\
\midrule
\multirow{7}{*}{mPLUG-Owl2}
& Vanilla & 91.67 & 83.33 & 175.00 \\
& VCD\textsuperscript{\dag} \cite{qu2025mvp} & --- & --- & 170.00 \\
& OPERA\textsuperscript{\dag} \cite{qu2025mvp} & --- & --- & 173.33 \\
& DoLa\textsuperscript{\dag} \cite{chen2025aid} & --- & --- & 167.00 \\
& HALC\textsuperscript{\dag} \cite{chen2025aid} & --- & --- & 167.00 \\
& VASparse\textsuperscript{\dag} \cite{chen2025aid} & --- & --- & 175.00 \\
& \textbf{SSAV} & \textbf{93.33} & \textbf{86.67} & \textbf{180.00} \\
\midrule
\multirow{6}{*}{MiniGPT-4 Vicuna-13B}
& Vanilla & 78.33 & 56.67 & 135.00 \\
& LRV-Instruction\textsuperscript{\dag} \cite{wu2024logiccheckgpt} & 83.33 & 66.67 & 150.00 \\
& SelfCheck\textsuperscript{\dag} \cite{wu2024logiccheckgpt} & 80.00 & 60.00 & 140.00 \\
& LURE\textsuperscript{\dag} \cite{wu2024logiccheckgpt} & 85.00 & 70.00 & 155.00 \\
& LogicCheckGPT \cite{wu2024logiccheckgpt} & 86.67 & 73.33 & 160.00 \\
& \textbf{SSAV} & \textbf{98.33} & \textbf{96.67} & \textbf{195.00} \\
\bottomrule
\end{tabular}
\end{table*}

SSAV achieves the highest MME Score in Table~\ref{tab:mme} for all three base models. For LLaVA-1.5-7B and mPLUG-Owl2, the score rises by 5 points from already strong baselines, with a larger gain in Acc+ than in Accuracy. Acc+ requires both paired questions for the same image to be answered correctly, so the pattern indicates that SSAV not only corrects individual existence judgments but also reduces inconsistency across paired Yes--No questions. For MiniGPT-4 Vicuna-13B, Accuracy increases from 78.33\% to 98.33\%, Acc+ from 56.67\% to 96.67\%, and the score from 135 to 195. The especially large Acc+ gain suggests that external semantic-spatial evidence is particularly effective at correcting systematic errors in paired questions. Together with the cross-source POPE evaluation, this result shows that SSAV's benefit for short-form object-existence decisions is not confined to one sampling protocol.

\begin{table*}[t]
\caption{Same-protocol comparison with representative methods on CHAIR. All results use 500 COCO images and a maximum generation length of 512 new tokens. Lower CHAIRs and CHAIRi are better. \textsuperscript{\dag} denotes results reported by subsequent studies using the same model backbone and generation protocol.}
\label{tab:chair}
\centering
\scriptsize
\setlength{\tabcolsep}{6pt}
\begin{tabular}{llcc}
\toprule
\textbf{Model} & \textbf{Method} & \textbf{CHAIRs $\downarrow$} & \textbf{CHAIRi $\downarrow$} \\
\midrule
\multirow{10}{*}{LLaVA-1.5-7B}
& Vanilla & 49.40 & 13.82 \\
& Nucleus\textsuperscript{\dag} \cite{dong2025inter} & 54.00 & 16.10 \\
& Nucleus+INTER \cite{dong2025inter} & 51.80 & 14.10 \\
& Beam\textsuperscript{\dag} \cite{dong2025inter} & 48.80 & 13.90 \\
& Beam+INTER \cite{dong2025inter} & 46.40 & 13.40 \\
& VCD\textsuperscript{\dag} \cite{dong2025inter} & 53.80 & 16.00 \\
& VCD+INTER \cite{dong2025inter} & 56.00 & 15.70 \\
& OPERA\textsuperscript{\dag} \cite{dong2025inter} & 45.40 & 13.80 \\
& OPERA+INTER \cite{dong2025inter} & 47.00 & 13.60 \\
& \textbf{SSAV} & \textbf{32.80} & \textbf{8.22} \\
\midrule
\multirow{10}{*}{mPLUG-Owl2}
& Vanilla & 60.00 & 17.32 \\
& Nucleus\textsuperscript{\dag} \cite{dong2025inter} & 60.80 & 20.10 \\
& Nucleus+INTER \cite{dong2025inter} & 59.40 & 19.30 \\
& Beam\textsuperscript{\dag} \cite{dong2025inter} & 56.40 & 17.90 \\
& Beam+INTER \cite{dong2025inter} & 53.40 & 17.20 \\
& VCD\textsuperscript{\dag} \cite{dong2025inter} & 62.80 & 20.50 \\
& VCD+INTER \cite{dong2025inter} & 60.40 & 20.50 \\
& OPERA\textsuperscript{\dag} \cite{dong2025inter} & 55.20 & 16.10 \\
& OPERA+INTER \cite{dong2025inter} & 52.50 & 15.90 \\
& \textbf{SSAV} & \textbf{36.00} & \textbf{10.14} \\
\midrule
\multirow{2}{*}{MiniGPT-4 Vicuna-13B}
& Vanilla & 32.00 & 9.23 \\
& \textbf{SSAV} & \textbf{19.40} & \textbf{5.98} \\
\bottomrule
\end{tabular}
\end{table*}

SSAV reduces both sentence-level and instance-level object hallucination for all three base models. On LLaVA-1.5-7B, CHAIRs decreases from 49.40\% to 32.80\% and CHAIRi from 13.82\% to 8.22\%. On mPLUG-Owl2, the metrics decrease from 60.00\% and 17.32\% to 36.00\% and 10.14\%; on MiniGPT-4 Vicuna-13B, they decrease from 32.00\% and 9.23\% to 19.40\% and 5.98\%. Their simultaneous reduction shows that SSAV reduces hallucination at both the sentence and object-instance levels, rather than merely removing a small number of concentrated errors. mPLUG-Owl2 has the highest baseline CHAIRs and the largest absolute reduction, indicating more room for object-level verification when the original captions hallucinate more severely. Continued gains on MiniGPT-4, whose baseline scores are lower, show that the mechanism does not depend on a high initial hallucination rate. Together with the matched-random deletion control in Table~\ref{tab:components}, these results indicate that the gain comes from evidence-based selection of claims rather than simple caption shortening.

\subsection{Ablation Studies}

To analyze the contribution of each SSAV component to object-hallucination mitigation, we conduct detailed ablations on LLaVA-1.5-7B under different evidence-construction and fusion settings. All variants use the same base-model outputs, OWL-ViT candidate responses, and evaluation samples. Query counts and evidence construction follow the definition of each variant, while all other implementation conditions remain unchanged; decision parameters remain fixed throughout evaluation.

\subsubsection{Ablation of Core Components}

The core-component ablation covers multi-prompt semantic support, QIRV spatial verification, and semantic-spatial fusion. For closed-set object-existence question answering, we compare five settings. Vanilla uses the original base-model decision without external visual evidence. Single Query uses only the top-1 response score to the class-name query. Prompt Mean averages top-1 scores across four semantically equivalent queries and retains only multi-prompt semantic support. Spatial QIRV uses only the spatial evidence composed of candidate persistence, spatial consistency, and within-query candidate dominance. Full SSAV fuses multi-prompt semantic support and QIRV with a geometric mean.

Each comparison addresses a specific question. Single Query versus Prompt Mean tests whether cross-query aggregation reduces response variation from one wording. Prompt Mean versus full SSAV measures the contribution of spatial-instance consistency beyond semantic response strength. Spatial QIRV versus full SSAV tests whether semantic support further improves a decision based only on region structure. Vanilla provides a common reference for the overall effect of each form of external evidence. Closed-set ablations report Accuracy and F1 on POPE; open-ended captioning compares Vanilla, Prompt Mean, Spatial QIRV, and full SSAV with CHAIRs and CHAIRi. We additionally include a Matched Random control for captioning, which randomly deletes the same number of object claims as full SSAV using the same editing procedure, testing whether improvements arise merely from the deletion count.

\begin{table}[t]
\caption{Ablation of core components on LLaVA-1.5-7B (\%). POPE results are reported on the GQA Adversarial subset. Lower CHAIRs and CHAIRi are better, and MME Score is the sum of Accuracy and Acc+. An em dash denotes an inapplicable setting; bold indicates the best result in each column.}
\label{tab:components}
\centering
\scriptsize
\setlength{\tabcolsep}{2.0pt}
\textit{(a) POPE and CHAIR}\\[2pt]
\begin{tabular}{lcccc}
\toprule
\textbf{Method} & \textbf{GQA Acc $\uparrow$} & \textbf{GQA F1 $\uparrow$} & \textbf{CHAIRs $\downarrow$} & \textbf{CHAIRi $\downarrow$} \\
\midrule
Vanilla & 77.73 & 80.54 & 49.40 & 13.82 \\
Single Query & 80.20 & 80.90 & --- & --- \\
Prompt Mean & 80.47 & 80.43 & 33.00 & 8.82 \\
Spatial QIRV & 81.23 & 81.62 & \textbf{32.80} & 8.84 \\
SSAV & \textbf{81.97} & \textbf{81.78} & \textbf{32.80} & \textbf{8.22} \\
Matched Random & --- & --- & 43.40 & 12.49 \\
\bottomrule
\end{tabular}
\vspace{4pt}

\textit{(b) MME-Existence}\\[2pt]
\begin{tabular}{lccc}
\toprule
\textbf{Method} & \textbf{Acc $\uparrow$} & \textbf{Acc+ $\uparrow$} & \textbf{Score $\uparrow$} \\
\midrule
Vanilla & 96.67 & 93.33 & 190.00 \\
Single Query & 96.67 & 93.33 & 190.00 \\
Prompt Mean & 96.67 & 93.33 & 190.00 \\
Spatial QIRV & 96.67 & 93.33 & 190.00 \\
SSAV & \textbf{98.33} & \textbf{96.67} & \textbf{195.00} \\
Matched Random & --- & --- & --- \\
\bottomrule
\end{tabular}
\end{table}

\textbf{Effect of multi-prompt aggregation and spatial evidence.} Single Query obtains 80.20\% Accuracy and 80.90\% F1 on GQA Adversarial. Prompt Mean slightly raises Accuracy to 80.47\% but lowers F1 to 80.43\%. Thus, aggregating queries reduces dependence on one wording but does not by itself guarantee simultaneous gains in both metrics; score averaging cannot determine whether high responses from different queries originate from the same instance. Spatial QIRV reaches 81.23\% Accuracy and 81.62\% F1, showing that cross-query regional correspondence contributes information beyond response strength. Full SSAV further reaches 81.97\% Accuracy and 81.78\% F1, supporting the complementarity of semantic support and spatial-instance consistency.

\textbf{Fusion gains in open-ended descriptions and paired judgments.} On CHAIR, Prompt Mean, Spatial QIRV, and full SSAV have similar CHAIRs, but full SSAV yields the lowest CHAIRi at 8.22\%, compared with 8.82\% and 8.84\% for Prompt Mean and Spatial QIRV. The fusion benefit therefore appears mainly in reducing hallucinated object instances after each individual branch has already removed the dominant sentence-level errors. On MME-Existence, Single Query, Prompt Mean, and Spatial QIRV all remain at 190, while only full SSAV reaches 195, further indicating that strict paired judgments benefit from joint support in response strength and regional consistency.

\subsubsection{QIRV Evidence Correspondence Analysis}

The core-component ablation establishes whether spatial evidence contributes to the final decision, but it does not determine whether QIRV's discriminative power derives from sample-specific spatial information rather than its score distribution or a dataset-level bias. We therefore progressively shuffle QIRV evidence to test whether correct correspondence between an object claim and its image region is necessary for effective discrimination.

The experiment is conducted on the POPE Adversarial subsets. QIRV evidence is shuffled only within each dataset at ratios of 0\%, 25\%, 50\%, 75\%, and 100\%. For ratios from 25\% to 100\%, we use random seeds 0, 1, 2, 3, and 4 and report the mean and standard deviation over five runs. Shuffling preserves the marginal QIRV score distribution within each dataset while progressively breaking its correspondence to individual image--claim samples. No task-decision threshold is used; we directly report ROC-AUC and PR-AUC, with PR-AUC computed as Average Precision.

As shown in Fig.~\ref{fig:sensitivity}(a), without shuffling, ROC-AUC and PR-AUC are 63.32\% and 67.29\%, respectively. At 25\% shuffling, they decrease to $59.95\%\pm0.56\%$ and $62.19\%\pm0.42\%$. At 50\% and 75\%, ROC-AUC further decreases to $56.67\%\pm0.45\%$ and $52.85\%\pm0.89\%$, while PR-AUC falls to $57.84\%\pm0.51\%$ and $53.52\%\pm0.77\%$. With fully shuffled QIRV evidence, ROC-AUC and PR-AUC are only $50.43\%\pm0.38\%$ and $50.78\%\pm0.42\%$, approaching random discrimination.

Both AUC metrics decrease continuously as the spatial-evidence correspondence is progressively destroyed. Relative to fully aligned QIRV, complete shuffling lowers ROC-AUC by 12.89 percentage points and PR-AUC by 16.51 points. Because the permutation preserves the marginal QIRV score distribution within each dataset, these results show that QIRV contains sample-specific spatial information. Its discrimination depends on correct matching between an object claim and the corresponding image region and cannot be explained by the overall score distribution alone.

\subsubsection{Sensitivity to Fusion Strength}

To analyze the effect of external-evidence calibration strength on closed-set object judgments, we vary $\lambda_{\mathrm{ssav}}\in\{0,1,2,3,4,5\}$ while holding all other settings fixed. $\lambda_{\mathrm{ssav}}=0$ corresponds to the original decision without the SSAV calibration term.

\begin{figure}[t]
\centering
\includegraphics[width=0.92\columnwidth]{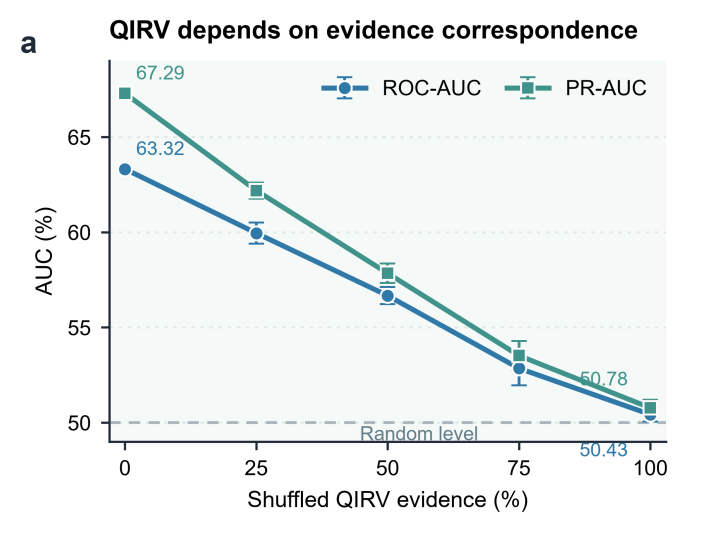}
\vspace{2pt}

\includegraphics[width=0.92\columnwidth]{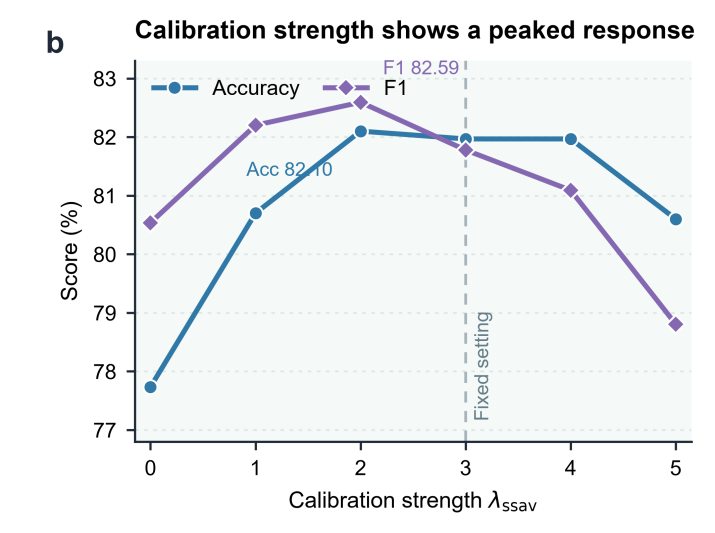}
\caption{QIRV evidence correspondence and calibration-strength sensitivity. (a) QIRV evidence is progressively shuffled within each of the three POPE Adversarial subsets while preserving the marginal score distribution. Points show macro-averaged ROC-AUC and PR-AUC; error bars denote the standard deviation over five random seeds, except for the deterministic 0\% setting. (b) Accuracy and F1 of LLaVA-1.5-7B as $\lambda_{\mathrm{SSAV}}$ varies, with all other parameters fixed.}
\label{fig:sensitivity}
\end{figure}

As shown in Fig.~\ref{fig:sensitivity}(b), increasing $\lambda_{\mathrm{ssav}}$ from 0 to 1 raises Accuracy from 77.73\% to 80.70\% and F1 from 80.54\% to 82.20\%. At $\lambda_{\mathrm{ssav}}=2$, they reach 82.10\% and 82.59\%. Over this range, the number of corrected errors grows from 0 to 240, while 109 new errors are introduced, yielding a net correction of 131. Moderate calibration therefore preferentially fixes original decisions lacking visual support. At larger strengths, corrected samples continue to increase, but harmed samples grow faster: the net correction is 127 at $\lambda_{\mathrm{ssav}}=3$ and 4 and drops to 86 at $\lambda_{\mathrm{ssav}}=5$. Recall simultaneously falls from 84.93\% at $\lambda_{\mathrm{ssav}}=2$ to 72.13\% at $\lambda_{\mathrm{ssav}}=5$. The rise-and-fall pattern in Accuracy and F1 therefore reflects over-calibration rather than failure of the external evidence: an excessive coefficient overwhelms the base-model margin and changes some true positives to No. Calibration strength must balance false-positive reduction against positive-class recall.

\begin{figure*}[!t]
\centering
\includegraphics[width=0.80\textwidth]{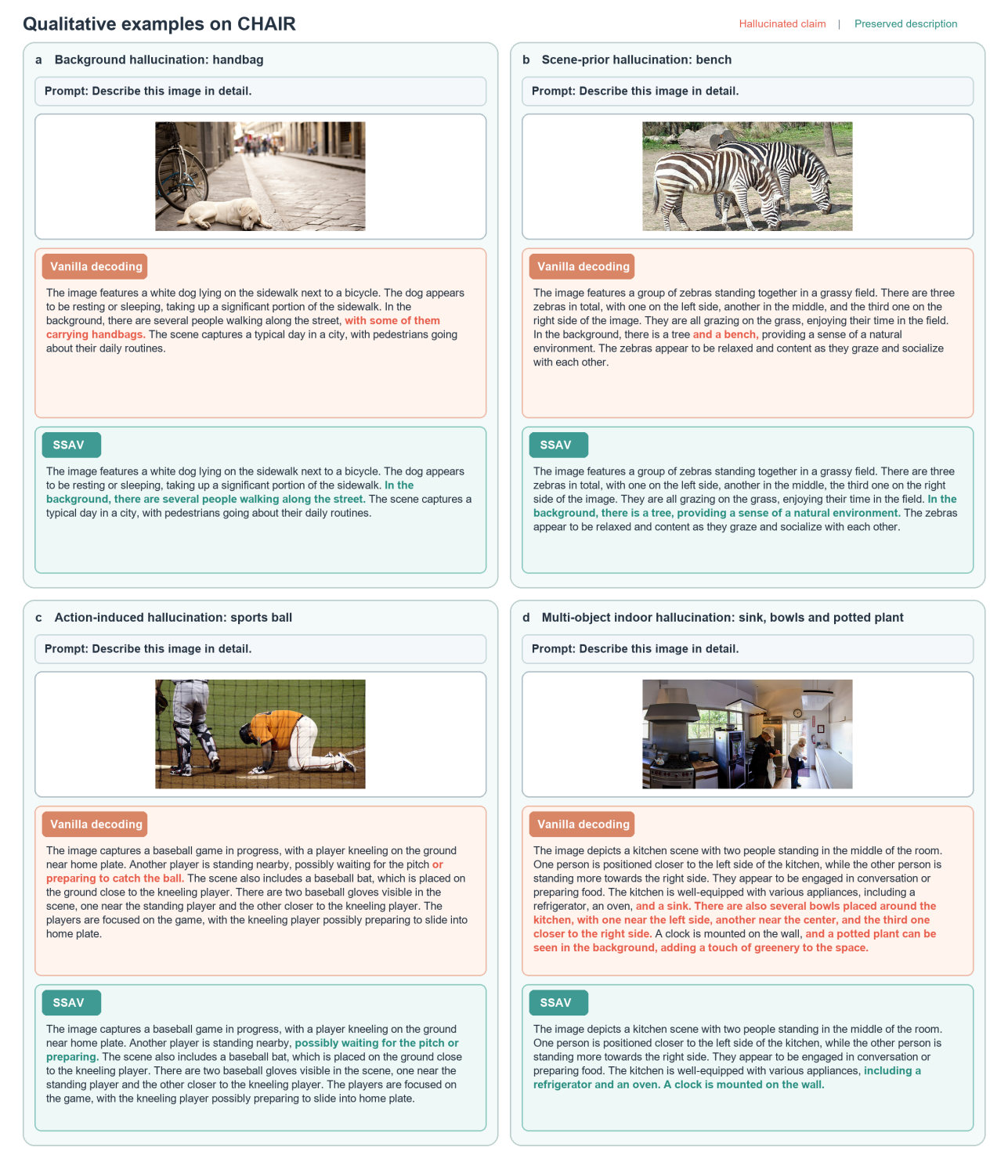}
\caption{Qualitative examples of SSAV on CHAIR. Hallucinated claims in the vanilla response are highlighted in red, and descriptions retained after SSAV filtering are highlighted in green. The four cases illustrate background-, scene-prior-, action-induced, and multi-object indoor hallucinations.}
\label{fig:qualitative}
\end{figure*}

Figure~\ref{fig:qualitative} illustrates claim-level revision by SSAV in four complementary scenarios. In case (a), the base model correctly identifies the dog, bicycle, and pedestrians but additionally states that some pedestrians carry handbags, an unsupported background-object claim. SSAV removes only the handbag phrase and preserves the remaining street-scene description. In case (b), a bench is linguistically plausible in a scene containing zebras, trees, and a natural environment, but the object is absent from the image. SSAV removes the bench while retaining the zebra count, positions, and background trees. In case (c), the action context of a baseball game induces the model to state that a player is preparing to catch a ball, although no visible sports-ball evidence is present. SSAV removes the corresponding object claim while retaining the visible players, bat, and gloves. Case (d) further demonstrates multi-object hallucination in a complex indoor scene: the base model mentions a sink, bowls, and a potted plant, whereas SSAV removes these unsupported objects while retaining the people, refrigerator, oven, and wall clock. The four cases respectively cover an accessory in a complex background, a co-occurring object induced by scene priors, an implicit object induced by action context, and multiple objects in an indoor scene. They show that semantic-spatial verification can revise local claims arising from different hallucination sources rather than truncating the description as a whole.

\FloatBarrier

\subsection{Design Choices and Robustness Analysis}

We compare fusion operators, IoU thresholds, and candidate-dominance formulations using 1,531 unambiguous claims from the complete CHAIR500 generations, including 1,158 present-object and 373 hallucinated claims; 38 claims with ambiguous category mappings are excluded. Confidence intervals use 1,000 bootstrap resamples clustered by image. For controlled ranking comparison, CHAIR metrics use a fixed rejection budget of 464 claims, matching calibrated full SSAV. These results therefore compare variants under a matched rejection budget rather than after independent calibration.

\subsubsection{Comparison of Fusion Operators}

\begin{table}[t]
\caption{Comparison of semantic-spatial fusion operators. Claim-level AUCs are computed on 1,531 unambiguous object claims; CHAIR results use a matched rejection budget of 464 claims.}
\label{tab:fusion}
\centering
\scriptsize
\resizebox{\columnwidth}{!}{%
\begin{tabular}{lccccc}
\toprule
\textbf{Fusion operator} & \textbf{ROC-AUC $\uparrow$} & \textbf{PR-AUC $\uparrow$} & \textbf{CHAIRs $\downarrow$} & \textbf{CHAIRi $\downarrow$} & \textbf{Recall $\uparrow$} \\
\midrule
Arithmetic mean & 81.23 & 93.15 & 34.40 & 8.91 & 74.86 \\
Geometric mean (SSAV) & 84.61 & 94.21 & 32.80 & 8.22 & 75.69 \\
Harmonic mean & 84.63 & 94.27 & 32.00 & 8.22 & 75.56 \\
Minimum & 84.62 & 94.33 & 31.80 & 8.34 & 75.56 \\
\bottomrule
\end{tabular}
}
\end{table}

The arithmetic mean obtains a claim-level ROC-AUC of 81.23\%, 3.39 points below the geometric mean, with a paired-bootstrap 95\% confidence interval of $[-4.59,-2.18]$. Its PR-AUC is 1.06 points lower, with an interval of $[-1.49,-0.65]$, showing that linear compensation can allow one branch to mask insufficient evidence from the other. The geometric mean, harmonic mean, and minimum provide similar discrimination, with no stable paired advantage over the geometric mean. We retain the geometric mean as a smooth, symmetric, and parameter-free soft conjunction of semantic and spatial evidence.

\subsubsection{Sensitivity to the IoU Threshold}

\begin{table}[t]
\caption{Sensitivity to the IoU threshold $\eta$. CHAIR results use the same matched rejection budget of 464 claims.}
\label{tab:iou}
\centering
\scriptsize
\resizebox{\columnwidth}{!}{%
\begin{tabular}{lccccc}
\toprule
\textbf{IoU threshold $\eta$} & \textbf{ROC-AUC $\uparrow$} & \textbf{PR-AUC $\uparrow$} & \textbf{CHAIRs $\downarrow$} & \textbf{CHAIRi $\downarrow$} & \textbf{Recall $\uparrow$} \\
\midrule
0.30 & 84.65 & 94.22 & 32.80 & 8.22 & 75.69 \\
0.40 & 84.66 & 94.22 & 32.80 & 8.22 & 75.62 \\
0.50 (default) & 84.61 & 94.21 & 32.80 & 8.22 & 75.69 \\
0.60 & 84.59 & 94.20 & 32.80 & 8.22 & 75.62 \\
0.70 & 84.57 & 94.20 & 32.80 & 8.22 & 75.62 \\
\bottomrule
\end{tabular}
}
\end{table}

As $\eta$ varies from 0.3 to 0.7, claim-level ROC-AUC and PR-AUC change by at most 0.09 and 0.02 percentage points, respectively. Under a matched rejection budget, CHAIRs and CHAIRi remain fixed at 32.80\% and 8.22\%, while Recall ranges only from 75.62\% to 75.69\%. These results demonstrate stable performance across the evaluated IoU thresholds, including the default $\eta=0.5$.

\subsubsection{Comparison of Candidate-Dominance Formulations}

\begin{table}[t]
\caption{Comparison of candidate-dominance formulations. CHAIR results use a matched rejection budget of 464 claims.}
\label{tab:dominance}
\centering
\scriptsize
\resizebox{\columnwidth}{!}{%
\begin{tabular}{lccccc}
\toprule
\textbf{Dominance formulation} & \textbf{ROC-AUC $\uparrow$} & \textbf{PR-AUC $\uparrow$} & \textbf{CHAIRs $\downarrow$} & \textbf{CHAIRi $\downarrow$} & \textbf{Recall $\uparrow$} \\
\midrule
Logit-sigmoid margin (SSAV) & 84.61 & 94.21 & 32.80 & 8.22 & 75.69 \\
Raw score gap & 82.91 & 93.55 & 32.40 & 8.37 & 76.01 \\
Relative score gap & 80.09 & 92.57 & 33.80 & 8.87 & 75.50 \\
Score ratio & 84.66 & 94.23 & 33.00 & 8.22 & 75.69 \\
Rank-only margin & 84.62 & 94.22 & 33.40 & 8.37 & 75.69 \\
\bottomrule
\end{tabular}
}
\end{table}

\begin{table}[t]
\caption{Inference efficiency on the complete CHAIR500 evaluation set. Wall-clock time excludes model loading and warm-up.}
\label{tab:efficiency}
\centering
\scriptsize
\resizebox{\columnwidth}{!}{%
\begin{tabular}{lccc}
\toprule
\textbf{Stage / Method} & \textbf{Total time} & \textbf{Latency per image} & \textbf{Peak reserved GPU memory} \\
\midrule
Vanilla LLaVA generation & 00:19:37 & 2.3550 s & 14.178 GB \\
SSAV external verification & 00:03:07 & 0.3744 s & 0.666 GB \\
Full SSAV pipeline & 00:22:45 & 2.7306 s & 14.178 GB \\
\bottomrule
\end{tabular}
}
\end{table}

The raw and relative score gaps reduce ROC-AUC to 82.91\% and 80.09\%, with paired-bootstrap differences of $-1.71$ points (95\% confidence interval $[-2.28,-1.20]$) and $-4.52$ points ($[-5.63,-3.48]$) from the logit-sigmoid margin. Thus, raw-score differences provide less stable claim-level ranking under the current detector-score distribution. The score ratio and rank-only margin achieve similar AUCs but no consistent advantage across the controlled CHAIR metrics. We retain the logit-sigmoid margin because it represents preferred-to-runner-up separation in a bounded form and provides a stable, interpretable trade-off among claim-level discrimination, CHAIR, and Recall.

\subsection{Efficiency Analysis}

Evaluated on the complete 500-image CHAIR set using an NVIDIA GeForce RTX 4090, Vanilla LLaVA requires 19 min 37 s, or 2.3550 s per image. The 1,569 generated object claims require 6,276 OWL-ViT calls. Verification, claim-level decisions, and text editing add 3 min 07 s, yielding a total of 22 min 45 s, or 2.7306 s per image. SSAV therefore requires 1.159$\times$ the Vanilla runtime, corresponding to a 15.9\% overhead. The verifier alone reserves 0.666 GB; because generation and verification run sequentially, the pipeline peak remains 14.178 GB. Measurements exclude model loading and warm-up.

\FloatBarrier

\section{Conclusion}

We investigate how an independent visual model can provide reliable external verification for object claims in MLLM responses. A single high response from an open-vocabulary detector can be affected by query wording and incidental local matches and therefore cannot be treated directly as sufficient evidence of object presence. By contrast, an object claim supported by real image content should maintain relatively stable responses under semantically equivalent queries and continue to point to the same visual instance. Based on this observation, we propose Semantic-Spatial Agreement Verification (SSAV). The method estimates semantic support through multi-prompt aggregation and introduces QIRV to jointly characterize candidate persistence, spatial overlap, and within-query candidate dominance, after which semantic and spatial evidence are fused into a unified object-level verification score. SSAV calibrates closed-set object-existence judgments and filters claims in open-ended image descriptions without updating the base model or the open-vocabulary detector. Experiments across multiple models and benchmarks show that SSAV effectively mitigates object hallucination in MLLMs.

SSAV's verification performance remains limited by the perceptual capability of the external detector and incurs additional inference cost. Because QIRV uses each query's preferred candidate region, small, occluded, or repeated-category objects may cause equivalent queries to localize to different real instances, weakening spatial evidence. Future work will examine alternative and multiple detectors to reduce detector-specific bias, extend Top-1 fixed-IoU verification to Top-$M$ candidates, adaptive spatial relations, and multi-instance matching, assess robustness to synonymous and multilingual queries, and reduce verification cost through batched queries, candidate caching, and shared visual features.

\bibliographystyle{IEEEtran}
\IEEEtriggeratref{50}
\bibliography{references_paper}

@inproceedings{an2025agla,
  title = {{{AGLA}}: {{Mitigating}} Object Hallucinations in Large Vision-Language Models with Assembly of Global and Local Attention},
  booktitle = {Proceedings of the {{IEEE}}/{{CVF}} Conference on Computer Vision and Pattern Recognition},
  author = {An, Wenbin and Tian, Feng and Leng, Sicong and Nie, Jiahao and Lin, Haonan and Wang, QianYing and Dai, Guang and Chen, Ping and Lu, Shijian},
  year = {2025}
}

@unpublished{chen2025aid,
  title = {Attention Hijackers: {{Detect}} and Disentangle Attention Hijacking in {{LVLMs}} for Hallucination Mitigation},
  author = {Chen, Beitao and Lyu, Xinyu and Gao, Lianli and Song, Jingkuan and Shen, Heng Tao},
  year = {2025},
  eprint = {2503.08216},
  eprinttype = {arXiv},
  eprintclass = {cs.CV},
  doi = {10.48550/arXiv.2503.08216},
  url = {https://arxiv.org/abs/2503.08216}
}

@article{daiInstructBLIPGeneralpurposeVisionLanguage2023,
  title = {{{InstructBLIP}}: {{Towards General-purpose Vision-Language Models}} with {{Instruction Tuning}}},
author = {Dai, Wenliang and Li, Junnan and Li, Dongxu and Tiong, Anthony and Zhao, Junqi and Wang, Weisheng and Li, Boyang and Fung, Pascale N. and Hoi, Steven},
  year = {2023},
  journal = {Advances in Neural Information Processing Systems},
  volume = {36},
  pages = {49250--49267},
langid = {english}
}

@inproceedings{dastmalchi2026cipher,
  title = {Fighting Hallucinations with Counterfactuals: {{Diffusion-guided}} Perturbations for {{LVLM}} Hallucination Suppression},
  booktitle = {Proceedings of the {{IEEE}}/{{CVF}} Conference on Computer Vision and Pattern Recognition},
  author = {Dastmalchi, Hamidreza and An, Aijun and Cheraghian, Ali and Barzamini, Hamed},
  year = {2026}
}

@inproceedings{dong2025inter,
  title = {{{INTER}}: {{Mitigating}} Hallucination in Large Vision-Language Models by Interaction Guidance Sampling},
  booktitle = {Proceedings of the {{IEEE}}/{{CVF}} International Conference on Computer Vision},
  author = {Dong, Xin and Dong, Shichao and Wang, Jin and Huang, Jing and Zhou, Li and Sun, Zenghui and Jing, Lihua and Lan, Jinsong and Zhu, Xiaoyong and Zheng, Bo},
  year = {2025},
  pages = {2534--2544},
  doi = {10.1109/ICCV51701.2025.00244}
}

@inproceedings{dosovitskiy2021image,
  title = {An Image Is Worth 16x16 Words: {{Transformers}} for Image Recognition at Scale},
  booktitle = {International Conference on Learning Representations},
  author = {Dosovitskiy, Alexey and Beyer, Lucas and Kolesnikov, Alexander and Weissenborn, Dirk and Zhai, Xiaohua and Unterthiner, Thomas and Dehghani, Mostafa and Minderer, Matthias and Heigold, Georg and Gelly, Sylvain and Uszkoreit, Jakob and Houlsby, Neil},
  year = {2021}
}

@inproceedings{duan2025truthprint,
  title = {{{TruthPrInt}}: {{Mitigating}} Large Vision-Language Models Object Hallucination via Latent Truthful-Guided Pre-Intervention},
  booktitle = {Proceedings of the {{IEEE}}/{{CVF}} International Conference on Computer Vision},
  author = {Duan, Jinhao and Kong, Fei and Cheng, Hao and Diffenderfer, James and Kailkhura, Bhavya and Sun, Lichao and Zhu, Xiaofeng and Shi, Xiaoshuang and Xu, Kaidi},
  year = {2025},
  doi = {10.1109/ICCV51701.2025.00692}
}

@inproceedings{fuMMEComprehensiveEvaluation2025,
  title = {{{MME}}: A Comprehensive Evaluation Benchmark for Multimodal Large Language Models},
  booktitle = {Advances in Neural Information Processing Systems: {{Datasets}} and Benchmarks Track},
  author = {Fu, Chaoyou and Chen, Peixian and Shen, Yunhang and Qin, Yulei and Zhang, Mengdan and Lin, Xu and Yang, Jinrui and Zheng, Xiawu and Li, Ke and Sun, Xing and Wu, Yunsheng and Ji, Rongrong},
  year = {2025}
}

@inproceedings{gu2022vild,
  title = {Open-Vocabulary Object Detection via Vision and Language Knowledge Distillation},
  booktitle = {International Conference on Learning Representations},
  author = {Gu, Xiuye and Lin, Tsung-Yi and Kuo, Weicheng and Cui, Yin},
  year = {2022}
}

@inproceedings{huang2024mllm,
  title = {{{MLLM}} Can See? {{Dynamic}} Correction Decoding for Hallucination Mitigation},
  booktitle = {International Conference on Learning Representations},
  author = {Wang, Chenxi and Chen, Xiang and Zhang, Ningyu and Tian, Bozhong and Xu, Haoming and Deng, Shumin and Chen, Huajun},
  year = {2025}
}

@inproceedings{huangOPERAAlleviatingHallucination2024,
  title = {{{OPERA}}: {{Alleviating Hallucination}} in {{Multi-Modal Large Language Models}} via {{Over-Trust Penalty}} and {{Retrospection-Allocation}}},
author = {Huang, Qidong and Dong, Xiaoyi and Zhang, Pan and Wang, Bin and He, Conghui and Wang, Jiaqi and Lin, Dahua and Zhang, Weiming and Yu, Nenghai},
  year = {2024},
  pages = {13418--13427},
booktitle = {Proceedings of the {{IEEE}}/{{CVF Conference}} on {{Computer Vision}} and {{Pattern Recognition}}},
  langid = {english}
}

@inproceedings{hudsonGQANewDataset2019,
  title = {{{GQA}}: {{A New Dataset}} for {{Real-World Visual Reasoning}} and {{Compositional Question Answering}}},
author = {Hudson, Drew A. and Manning, Christopher D.},
  year = {2019},
  pages = {6700--6709},
booktitle = {Proceedings of the {{IEEE}}/{{CVF Conference}} on {{Computer Vision}} and {{Pattern Recognition}}}
}

@inproceedings{jiang2025devils,
  title = {Devils in Middle Layers of Large Vision-Language Models: {{Interpreting}}, Detecting and Mitigating Object Hallucinations via Attention Lens},
  booktitle = {Proceedings of the {{IEEE}}/{{CVF}} Conference on Computer Vision and Pattern Recognition},
  author = {Jiang, Zhangqi and Chen, Junkai and Zhu, Beier and Luo, Tingjin and Shen, Yankun and Yang, Xu},
  year = {2025},
  pages = {25004--25014},
  doi = {10.1109/CVPR52734.2025.02328}
}

@inproceedings{jiang2026kvsmooth,
  title = {{{KVSmooth}}: {{Mitigating}} Hallucination in Multi-Modal Large Language Models through Key-Value Smoothing},
  booktitle = {Proceedings of the {{IEEE}}/{{CVF}} Conference on Computer Vision and Pattern Recognition},
  author = {Jiang, Siyu and Chen, Feiyang and Zhang, Xiaojin and He, Kun},
  year = {2026}
}

@inproceedings{leng2024mitigating,
  title = {Mitigating Object Hallucinations in Large Vision-Language Models through Visual Contrastive Decoding},
  booktitle = {Proceedings of the {{IEEE}}/{{CVF}} Conference on Computer Vision and Pattern Recognition ({{CVPR}})},
  author = {Leng, Sicong and Zhang, Hang and Chen, Guanzheng and Li, Xin and Lu, Shijian and Miao, Chunyan and Bing, Lidong},
  year = {2024},
  pages = {13872--13882}
}

@inproceedings{li2022glip,
  title = {Grounded Language-Image Pre-Training},
  booktitle = {Proceedings of the {{IEEE}}/{{CVF}} Conference on Computer Vision and Pattern Recognition},
  author = {Li, Liunian Harold and Zhang, Pengchuan and Zhang, Haotian and Yang, Jianwei and Li, Chunyuan and Zhong, Yiwu and Wang, Lijuan and Yuan, Lu and Zhang, Lei and Hwang, Jenq-Neng and Chang, Kai-Wei and Gao, Jianfeng},
  year = {2022},
  pages = {10965--10975},
  doi = {10.1109/CVPR52688.2022.01069}
}

@inproceedings{li2023blip2,
  title = {{{BLIP-2}}: {{Bootstrapping}} Language-Image Pre-Training with Frozen Image Encoders and Large Language Models},
  booktitle = {Proceedings of the 40th International Conference on Machine Learning},
  author = {Li, Junnan and Li, Dongxu and Savarese, Silvio and Hoi, Steven C. H.},
  year = {2023},
  series = {Proceedings of Machine Learning Research},
  volume = {202},
  pages = {19730--19742},
  publisher = {PMLR}
}

@inproceedings{li2024vlfeedback,
  title = {{{VLFeedback}}: A Large-Scale {{AI}} Feedback Dataset for Large Vision-Language Models Alignment},
  booktitle = {Proceedings of the 2024 Conference on Empirical Methods in Natural Language Processing},
  author = {Li, Lei and Xie, Zhihui and Li, Mukai and Chen, Shunian and Wang, Peiyi and Chen, Liang and Yang, Yazheng and Wang, Benyou and Kong, Lingpeng and Liu, Qi},
  year = {2024},
  pages = {6227--6246},
  publisher = {Association for Computational Linguistics},
  doi = {10.18653/v1/2024.emnlp-main.358}
}

@inproceedings{li2026cmac,
  title = {Cross-Modal Attention Calibration for {{LVLM}} Hallucination Mitigation},
  booktitle = {Proceedings of the {{IEEE}}/{{CVF}} Conference on Computer Vision and Pattern Recognition},
  author = {Li, Jiaming and Zhang, Jiacheng and Jie, Zequn and Ma, Lin and Li, Ming and Luo, Xiaonan and Li, Guanbin},
  year = {2026},
  pages = {40186--40196}
}

@inproceedings{liEvaluatingObjectHallucination2023,
  title = {Evaluating {{Object Hallucination}} in {{Large Vision-Language Models}}},
  booktitle = {Proceedings of the 2023 {{Conference}} on {{Empirical Methods}} in {{Natural Language Processing}}},
  author = {Li, Yifan and Du, Yifan and Zhou, Kun and Wang, Jinpeng and Zhao, Xin and Wen, Ji-Rong},
  editor = {Bouamor, Houda and Pino, Juan and Bali, Kalika},
  year = {2023},
  pages = {292--305},
  publisher = {Association for Computational Linguistics},
  location = {Singapore},
  doi = {10.18653/v1/2023.emnlp-main.20},
eventtitle = {{{EMNLP}} 2023}
}

@inproceedings{linMicrosoftCOCOCommon2014,
  title = {Microsoft {{COCO}}: {{Common Objects}} in {{Context}}},
booktitle = {Computer {{Vision}} – {{ECCV}} 2014},
  author = {Lin, Tsung-Yi and Maire, Michael and Belongie, Serge and Hays, James and Perona, Pietro and Ramanan, Deva and Dollár, Piotr and Zitnick, C. Lawrence},
  editor = {Fleet, David and Pajdla, Tomas and Schiele, Bernt and Tuytelaars, Tinne},
  year = {2014},
  pages = {740--755},
  publisher = {Springer International Publishing},
  location = {Cham},
  doi = {10.1007/978-3-319-10602-1_48},
isbn = {978-3-319-10602-1}
}

@inproceedings{liu2024groundingdino,
  title = {Grounding {{DINO}}: {{Marrying DINO}} with Grounded Pre-Training for Open-Set Object Detection},
  booktitle = {European Conference on Computer Vision},
  author = {Liu, Shilong and Zeng, Zhaoyang and Ren, Tianhe and Li, Feng and Zhang, Hao and Yang, Jie and Jiang, Qing and Li, Chunyuan and Yang, Jianwei and Su, Hang and Zhu, Jun and Zhang, Lei},
  year = {2024},
  pages = {38--55},
  publisher = {Springer},
  doi = {10.1007/978-3-031-72970-6_3}
}

@inproceedings{liu2024mitigating,
  title = {Mitigating Hallucination in Large Multi-Modal Models via Robust Instruction Tuning},
  booktitle = {The Twelfth International Conference on Learning Representations},
  author = {Liu, Fuxiao and Lin, Kevin and Li, Linjie and Wang, Jianfeng and Yacoob, Yaser and Wang, Lijuan},
  year = {2024}
}

@inproceedings{liu2024paying,
  title = {Paying More Attention to Image: A Training-Free Method for Alleviating Hallucination in Lvlms},
  booktitle = {European Conference on Computer Vision ({{ECCV}})},
  author = {Liu, Shi and Zheng, Kecheng and Chen, Wei},
  year = {2024},
  publisher = {Springer}
}

@unpublished{liu2025energy,
  title = {Energy-Guided Decoding for Object Hallucination Mitigation},
  author = {Liu, Xixi and Deng, Ailin and Zach, Christopher},
  year = {2025},
  eprint = {2507.07731},
  eprinttype = {arXiv},
  eprintclass = {cs.CV},
  url = {https://arxiv.org/abs/2507.07731}
}

@inproceedings{liuImprovedBaselinesVisual2024,
  title = {Improved {{Baselines}} with {{Visual Instruction Tuning}}},
  booktitle = {2024 {{IEEE}}/{{CVF Conference}} on {{Computer Vision}} and {{Pattern Recognition}} ({{CVPR}})},
  author = {Liu, Haotian and Li, Chunyuan and Li, Yuheng and Lee, Yong Jae},
  year = {2024},
  pages = {26286--26296},
  publisher = {IEEE},
  location = {Seattle, WA, USA},
  doi = {10.1109/CVPR52733.2024.02484},
isbn = {979-8-3503-5300-6}
}

@inproceedings{minderer2022owlvit,
  title = {Simple Open-Vocabulary Object Detection with Vision Transformers},
  booktitle = {European Conference on Computer Vision},
  author = {Minderer, Matthias and Gritsenko, Alexey and Stone, Austin and Neumann, Maxim and Weissenborn, Dirk and Dosovitskiy, Alexey and Mahendran, Aravindh and Arnab, Anurag and Dehghani, Mostafa and Shen, Zhuoran and Wang, Xiao and Zhai, Xiaohua and Kipf, Thomas and Houlsby, Neil},
  year = {2022},
  pages = {728--755},
  publisher = {Springer}
}

@inproceedings{NEURIPS2023_6dcf277e,
  title = {Visual Instruction Tuning},
  booktitle = {Advances in Neural Information Processing Systems},
  author = {Liu, Haotian and Li, Chunyuan and Wu, Qingyang and Lee, Yong Jae},
  editor = {Oh, A. and Naumann, T. and Globerson, A. and Saenko, K. and Hardt, M. and Levine, S.},
  year = {2023},
  volume = {36},
  pages = {34892--34916},
  publisher = {Curran Associates, Inc.}
}

@inproceedings{qu2025mvp,
  title = {Look, Compare, Decide: {{Alleviating}} Hallucination in Large Vision-Language Models via Multi-View Multi-Path Reasoning},
  booktitle = {Proceedings of the 31st International Conference on Computational Linguistics},
  author = {Qu, Xiaoye and Sun, Jiashuo and Wei, Wei and Liu, Daizong and Dong, Jianfeng and Cheng, Yu},
  year = {2025},
  pages = {4428--4441},
  publisher = {Association for Computational Linguistics},
  location = {Abu Dhabi, UAE}
}

@inproceedings{radfordLearningTransferableVisual2021,
  title = {Learning {{Transferable Visual Models From Natural Language Supervision}}},
  booktitle = {Proceedings of the 38th {{International Conference}} on {{Machine Learning}}},
  author = {Radford, Alec and Kim, Jong Wook and Hallacy, Chris and Ramesh, Aditya and Goh, Gabriel and Agarwal, Sandhini and Sastry, Girish and Askell, Amanda and Mishkin, Pamela and Clark, Jack and Krueger, Gretchen and Sutskever, Ilya},
  year = {2021},
  pages = {8748--8763},
  publisher = {PMLR},
langid = {english}
}

@inproceedings{rohrbachObjectHallucinationImage2018,
  title = {Object {{Hallucination}} in {{Image Captioning}}},
  booktitle = {Proceedings of the 2018 {{Conference}} on {{Empirical Methods}} in {{Natural Language Processing}}},
  author = {Rohrbach, Anna and Hendricks, Lisa Anne and Burns, Kaylee and Darrell, Trevor and Saenko, Kate},
  editor = {Riloff, Ellen and Chiang, David and Hockenmaier, Julia and Tsujii, Jun'ichi},
  year = {2018},
  pages = {4035--4045},
  publisher = {Association for Computational Linguistics},
  location = {Brussels, Belgium},
  doi = {10.18653/v1/D18-1437},
eventtitle = {{{EMNLP}} 2018}
}

@inproceedings{schwenkAOKVQABenchmarkVisual2022,
  title = {A-{{OKVQA}}: {{A Benchmark}} for {{Visual Question Answering Using World Knowledge}}},
booktitle = {Computer {{Vision}} – {{ECCV}} 2022},
  author = {Schwenk, Dustin and Khandelwal, Apoorv and Clark, Christopher and Marino, Kenneth and Mottaghi, Roozbeh},
  editor = {Avidan, Shai and Brostow, Gabriel and Cissé, Moustapha and Farinella, Giovanni Maria and Hassner, Tal},
  year = {2022},
  pages = {146--162},
  publisher = {Springer Nature Switzerland},
  location = {Cham},
  doi = {10.1007/978-3-031-20074-8_9},
isbn = {978-3-031-20074-8},
  langid = {english}
}

@inproceedings{seo2025epistemic,
  title = {On Epistemic Uncertainty of Visual Tokens for Object Hallucinations in Large Vision-Language Models},
  booktitle = {Advances in Neural Information Processing Systems},
  author = {Seo, Hoigi and Kang, Dong Un and Cho, Hyunjin and Lee, Joohoon and Chun, Se Young},
  year = {2025}
}

@inproceedings{wu2024logiccheckgpt,
  title = {Logical Closed Loop: {{Uncovering}} Object Hallucinations in Large Vision-Language Models},
  booktitle = {Findings of the Association for Computational Linguistics: {{ACL}} 2024},
  author = {Wu, Junfei and Liu, Qiang and Wang, Ding and Zhang, Jinghao and Wu, Shu and Wang, Liang and Tan, Tieniu},
  year = {2024},
  pages = {6944--6962},
  publisher = {Association for Computational Linguistics},
  location = {Bangkok, Thailand},
  doi = {10.18653/v1/2024.findings-acl.414}
}

@unpublished{wu2026revis,
  title = {{{REVIS}}: {{Sparse}} Latent Steering to Mitigate Object Hallucination in Large Vision-Language Models},
  author = {Wu, Jialin and Shi, Wei and Shen, Han and Qi, Peigui and Tang, Kunsheng and Huang, Zhicong and Wang, Binghao and Yang, Zhou},
  year = {2026},
  eprint = {2602.11824},
  eprinttype = {arXiv},
  url = {https://arxiv.org/abs/2602.11824}
}

@inproceedings{xing2024cca,
  title = {Mitigating Object Hallucination via Concentric Causal Attention},
  booktitle = {Advances in Neural Information Processing Systems},
  author = {Xing, Yun and Li, Yiheng and Laptev, Ivan and Lu, Shijian},
  year = {2024}
}

@inproceedings{yeMPLUGOwl2RevolutionizingMultimodal2024,
  title = {{{mPLUG-Owl2}}: {{Revolutionizing Multi-modal Large Language Model}} with {{Modality Collaboration}}},
author = {Ye, Qinghao and Xu, Haiyang and Ye, Jiabo and Yan, Ming and Hu, Anwen and Liu, Haowei and Qian, Qi and Zhang, Ji and Huang, Fei},
  year = {2024},
  pages = {13040--13051},
booktitle = {Proceedings of the {{IEEE}}/{{CVF Conference}} on {{Computer Vision}} and {{Pattern Recognition}}},
  langid = {english}
}

@unpublished{yin2023woodpecker,
  title = {Woodpecker: {{Hallucination}} Correction for Multimodal Large Language Models},
  author = {Yin, Shukang and Fu, Chaoyou and Zhao, Sirui and Xu, Tong and Wang, Hao and Sui, Dianbo and Shen, Yunhang and Li, Ke and Sun, Xing and Chen, Enhong},
  year = {2023},
  eprint = {2310.16045},
  eprinttype = {arXiv},
  url = {https://arxiv.org/abs/2310.16045}
}

@inproceedings{yin2025clearsight,
  title = {{{ClearSight}}: {{Visual}} Signal Enhancement for Object Hallucination Mitigation in Multimodal Large Language Models},
  booktitle = {Proceedings of the {{IEEE}}/{{CVF}} Conference on Computer Vision and Pattern Recognition ({{CVPR}})},
  author = {Yin, Hao and Si, Guangzong and Wang, Zilei},
  year = {2025}
}

@inproceedings{zhang2025degf,
  title = {Self-Correcting Decoding with Generative Feedback for Mitigating Hallucinations in Large Vision-Language Models},
  booktitle = {International Conference on Learning Representations},
  author = {Zhang, Ce and Wan, Zifu and Kan, Zhehan and Ma, Martin Q. and Stepputtis, Simon and Ramanan, Deva and Salakhutdinov, Russ and Morency, Louis-Philippe and Sycara, Katia P. and Xie, Yaqi},
  year = {2025}
}

@inproceedings{zhang2026pti,
  title = {Prefill-Time Intervention for Mitigating Hallucination in Large Vision-Language Models},
  booktitle = {Proceedings of the {{IEEE}}/{{CVF}} Conference on Computer Vision and Pattern Recognition},
  author = {Zhang, Chengsheng and Sun, Chenghao and Jiang, Xinyan and Li, Wei and Tian, Xinmei},
  year = {2026}
}

@inproceedings{zhuang2025vasparse,
  title = {{{VASparse}}: {{Towards}} Efficient Visual Hallucination Mitigation via Visual-Aware Token Sparsification},
  booktitle = {Proceedings of the {{IEEE}}/{{CVF}} Conference on Computer Vision and Pattern Recognition},
  author = {Zhuang, Xianwei and Zhu, Zhihong and Xie, Yuxin and Liang, Liming and Zou, Yuexian},
  year = {2025}
}

@inproceedings{zhuMiniGPT4EnhancingVisionLanguage2023,
  title = {{{MiniGPT-4}}: {{Enhancing Vision-Language Understanding}} with {{Advanced Large Language Models}}},
author = {Zhu, Deyao and Chen, Jun and Shen, Xiaoqian and Li, Xiang and Elhoseiny, Mohamed},
  year = {2023},
booktitle = {The {{Twelfth International Conference}} on {{Learning Representations}}},
  langid = {english}
}

@inproceedings{zouLookTwiceYou2025,
  title = {Look {{Twice Before You Answer}}: {{Memory-Space Visual Retracing}} for {{Hallucination Mitigation}} in {{Multimodal Large Language Models}}},
author = {Zou, Xin and Wang, Yizhou and Yan, Yibo and Lyu, Yuanhuiyi and Zheng, Kening and Huang, Sirui and Chen, Junkai and Jiang, Peijie and Liu, Jia and Tang, Chang and Hu, Xuming},
  year = {2025},
booktitle = {Forty-Second {{International Conference}} on {{Machine Learning}}},
  langid = {english}
}

@inproceedings{mao2026perceptionMagnifier,
  title = {Through the Magnifying Glass: Adaptive Perception Magnification for Hallucination-Free {VLM} Decoding},
  booktitle = {Proceedings of the 64th Annual Meeting of the Association for Computational Linguistics (Volume 1: Long Papers)},
  author = {Mao, Shunqi and Zhang, Chaoyi and Cai, Weidong},
  year = {2026},
  month = jul,
  pages = {44480--44501},
  publisher = {Association for Computational Linguistics},
  address = {San Diego, California, United States},
  doi = {10.18653/v1/2026.acl-long.2059}
}

@inproceedings{tang2025blindHallucination,
  title = {{``This Is My Fault'', Really? Understanding Blind and Low-Vision People's Perception of Hallucination in Large Vision Language Models}},
  booktitle = {Proceedings of the 38th Annual {ACM} Symposium on User Interface Software and Technology},
  author = {Tang, Yilin and Fang, Yuyang and Wang, Tianle and Sun, Lingyun and Chen, Liuqing},
  year = {2025},
  pages = {1--20},
  publisher = {Association for Computing Machinery},
  address = {New York, NY, USA},
  doi = {10.1145/3746059.3747597},
  articleno = {44}
}

@inproceedings{zhangLowRankPromptAdaptation,
  title = {Low-Rank Prompt Adaptation for Open-Vocabulary Object Detection},
  booktitle = {Proceedings of the {IEEE}/{CVF} International Conference on Computer Vision Workshops},
  author = {Zhang, Zekun and Truong, Vu Quang and Hoai, Minh},
  year = {2025},
  pages = {4263--4274}
}

@inproceedings{chen2024halc,
  author = {Chen, Zhaorun and Zhao, Zhuokai and Luo, Hongyin and Yao, Huaxiu and Li, Bo and Zhou, Jiawei},
  title = {{HALC}: Object Hallucination Reduction via Adaptive Focal-Contrast Decoding},
  booktitle = {Proceedings of the 41st International Conference on Machine Learning},
  series = {Proceedings of Machine Learning Research},
  volume = {235},
  pages = {7824--7846},
  year = {2024},
  publisher = {PMLR},
  url = {https://proceedings.mlr.press/v235/chen24bi.html}
}

@inproceedings{park2025convis,
  author = {Park, Yeji and Lee, Deokyeong and Choe, Junsuk and Chang, Buru},
  title = {{ConVis}: Contrastive Decoding with Hallucination Visualization for Mitigating Hallucinations in Multimodal Large Language Models},
  booktitle = {Proceedings of the AAAI Conference on Artificial Intelligence},
  volume = {39},
  number = {6},
  pages = {6434--6442},
  year = {2025},
  doi = {10.1609/aaai.v39i6.32689}
}

@inproceedings{wang2024mitigating,
  author = {Wang, Xintong and Pan, Jingheng and Ding, Liang and Biemann, Chris},
  title = {Mitigating Hallucinations in Large Vision-Language Models with Instruction Contrastive Decoding},
  booktitle = {Findings of the Association for Computational Linguistics: ACL 2024},
  pages = {15840--15853},
  year = {2024},
  doi = {10.18653/v1/2024.findings-acl.937}
}

@inproceedings{huo2025self,
  author = {Huo, Fushuo and Xu, Wenchao and Zhang, Zhong and Wang, Haozhao and Chen, Zhicheng and Zhao, Peilin},
  title = {Self-Introspective Decoding: Alleviating Hallucinations for Large Vision-Language Models},
  booktitle = {The Thirteenth International Conference on Learning Representations},
  year = {2025},
  url = {https://openreview.net/forum?id=rsZwwjYHuD}
}

@inproceedings{zhou2024lure,
  author = {Zhou, Yiyang and Cui, Chenhang and Yoon, Jaehong and Zhang, Linjun and Deng, Zhun and Finn, Chelsea and Bansal, Mohit and Yao, Huaxiu},
  title = {Analyzing and Mitigating Object Hallucination in Large Vision-Language Models},
  booktitle = {The Twelfth International Conference on Learning Representations},
  year = {2024},
  url = {https://openreview.net/forum?id=oZDJKTlOUe}
}

@inproceedings{duLearningPromptOpenVocabulary2022,
  author = {Du, Yu and Wei, Fangyun and Zhang, Zihe and Shi, Miaojing and Gao, Yue and Li, Guoqi},
  title = {Learning to Prompt for Open-Vocabulary Object Detection with Vision-Language Model},
  booktitle = {Proceedings of the IEEE/CVF Conference on Computer Vision and Pattern Recognition},
  pages = {14064--14073},
  year = {2022},
  doi = {10.1109/CVPR52688.2022.01369}
}

@inproceedings{zhong2022regionclip,
  author = {Zhong, Yiwu and Yang, Jianwei and Zhang, Pengchuan and Li, Chunyuan and Codella, Noel and Li, Liunian Harold and Zhou, Luowei and Dai, Xiyang and Yuan, Lu and Li, Yin and Gao, Jianfeng},
  title = {{RegionCLIP}: Region-Based Language-Image Pretraining},
  booktitle = {Proceedings of the IEEE/CVF Conference on Computer Vision and Pattern Recognition},
  pages = {16772--16782},
  year = {2022},
  doi = {10.1109/CVPR52688.2022.01629}
}

@inproceedings{zhou2022detic,
  author = {Zhou, Xingyi and Girdhar, Rohit and Joulin, Armand and Krahenbuhl, Philipp and Misra, Ishan},
  title = {Detecting Twenty-Thousand Classes Using Image-Level Supervision},
  booktitle = {Computer Vision -- ECCV 2022},
  pages = {350--368},
  year = {2022},
  publisher = {Springer},
  doi = {10.1007/978-3-031-20077-9_21}
}

@inproceedings{kuo2023fvlm,
  author = {Kuo, Weicheng and Cui, Yin and Gu, Xiuye and Piergiovanni, A. J. and Angelova, Anelia},
  title = {{F-VLM}: Open-Vocabulary Object Detection upon Frozen Vision and Language Models},
  booktitle = {The Eleventh International Conference on Learning Representations},
  year = {2023},
  url = {https://arxiv.org/abs/2209.15639}
}

@inproceedings{minderer2023scaling,
  author = {Minderer, Matthias and Gritsenko, Alexey A. and Houlsby, Neil},
  title = {Scaling Open-Vocabulary Object Detection},
  booktitle = {Advances in Neural Information Processing Systems},
  volume = {36},
  year = {2023},
  url = {https://openreview.net/forum?id=mQPNcBWjGc}
}

@article{wu2024openvocabulary,
  author = {Wu, Jianzong and Li, Xiangtai and Xu, Shilin and Yuan, Haobo and Ding, Henghui and Yang, Yibo and Li, Xia and Zhang, Jiangning and Tong, Yunhai and Jiang, Xudong and Ghanem, Bernard and Tao, Dacheng},
  title = {Towards Open Vocabulary Learning: A Survey},
  journal = {IEEE Transactions on Pattern Analysis and Machine Intelligence},
  volume = {46},
  number = {7},
  pages = {5092--5113},
  year = {2024},
  doi = {10.1109/TPAMI.2024.3361862}
}

\end{document}